\documentclass[]{style}

\usepackage{amsmath}
\usepackage{amssymb}
\usepackage{mathtools}
\usepackage{multirow}
\usepackage{makecell}
\usepackage{tabularx}
\usepackage{enumitem}
\usepackage{wrapfig}
\usepackage{pifont}
\usepackage{url}

\definecolor{RowBlue}{HTML}{E9F2FB}
\definecolor{RowRed}{HTML}{F9EAEA}
\newcommand{\cmark}{\textcolor{green!60!black}{\checkmark}}
\newcommand{\xmark}{\textcolor{red!70!black}{\ding{55}}}

\crefname{figure}{Fig.}{Figs.}
\renewcommand{\emph}[1]{\textit{#1}}

\usepackage{lmodern}\renewcommand{\sfdefault}{sans}

\usepackage{amsmath,amsfonts,bm}

\def\eqref#1{equation~\ref{#1}}

\def\1{\bm{1}}

\DeclareMathAlphabet{\mathsfit}{\encodingdefault}{\sfdefault}{m}{sl}
\SetMathAlphabet{\mathsfit}{bold}{\encodingdefault}{\sfdefault}{bx}{n}

\usepackage{booktabs}
\usepackage{array}
\usepackage{xspace}
\usepackage{adjustbox}
\usepackage{multirow}
\usepackage{colortbl}
\usepackage{listings}
\newcolumntype{L}[1]{>{\raggedright\arraybackslash}p{#1}}

\newif\ifwidelayout
\definecolor{pkulogo}{HTML}{9A0000}
\definecolor{codegray}{HTML}{6E6E6E}
\lstdefinestyle{prog}{
  language=Python, basicstyle=\ttfamily\scriptsize, keywordstyle=\color{pkulogo}\bfseries,
  commentstyle=\color{codegray}\itshape, stringstyle=\color{black!75}, showstringspaces=false,
  frame=single, rulecolor=\color{black!25}, framesep=3pt, xleftmargin=4pt, xrightmargin=4pt,
  columns=fullflexible, keepspaces=true, aboveskip=4pt, belowskip=2pt}

\usepackage{newfloat}
\usepackage{tcolorbox}
\DeclareFloatingEnvironment[name=Algorithm,placement=t]{algorithm}
\newtcolorbox{algobox}{colback=RowRed!45!white,colframe=pkulogo!55,boxrule=0.5pt,arc=2pt,
  left=5pt,right=5pt,top=3pt,bottom=3pt,boxsep=1pt}
\newcommand{\algkw}[1]{\textbf{#1}}
\newcommand{\algcmt}[1]{{\color{black!50}$\triangleright$\,#1}}
\newcommand{\algstage}[1]{{\sffamily\bfseries\color{pkulogo}#1}}
\newcommand{\ours}{PyRUA-Dex\xspace}
\newcommand{\stageone}{Stage~1\xspace}
\newcommand{\stagetwo}{Stage~2\xspace}

\newcommand{\papertitle}{\ours: Learning Programmatic Policies for Dexterous Bimanual Manipulation with Coding Agents}

\newcommand{\leadTwo}{15}
\newcommand{\tokOne}{11.7}
\newcommand{\tokTwo}{6.6}
\newcommand{\hrsOne}{615}
\newcommand{\hrsTwo}{362}

\titleformat*{\paragraph}{\bfseries}
\widelayouttrue
\usepackage{needspace}

\newcommand{\teasercaption}{\textbf{\ours at a glance.} Center: our real-world platform playing the piano (a video
frame, cut out), with tactile sensing at the fingertips. \textbf{(a)}~Coding agents learn a task's program in
simulation in two stages, exploration and refinement; the frozen program is deployed in the real world (sim-to-real),
and real-world feedback updates the simulation (real-to-sim; Section~\ref{sec:method}). \textbf{(b)}~DexJoCo success
rate with the objects randomized (rand-obj) and with cameras, lighting and textures randomized as well (rand-full;
Section~\ref{sec:res-dexjoco}). \textbf{(c)}~Tasks learned for the real-world platform: validation layouts completed
in simulation, of 32 (Section~\ref{sec:real}). \textbf{(d)}~The nine best of the thirteen four-channel entries on the
Bench2Dex leaderboard, where \ours holds the SOTA (Section~\ref{sec:res-main}). \textbf{(e)}~The hands the programs
drive, rendered from their models, with the joints driven per hand (DoF) and whether a program receives tactile
sensing. \textbf{(f)}~The tasks of the three settings in five classes, with the number of tasks per setting and one
picture each from simulation and from the real world.}

\title{\papertitle}

\secondleftlogo[8.2mm]{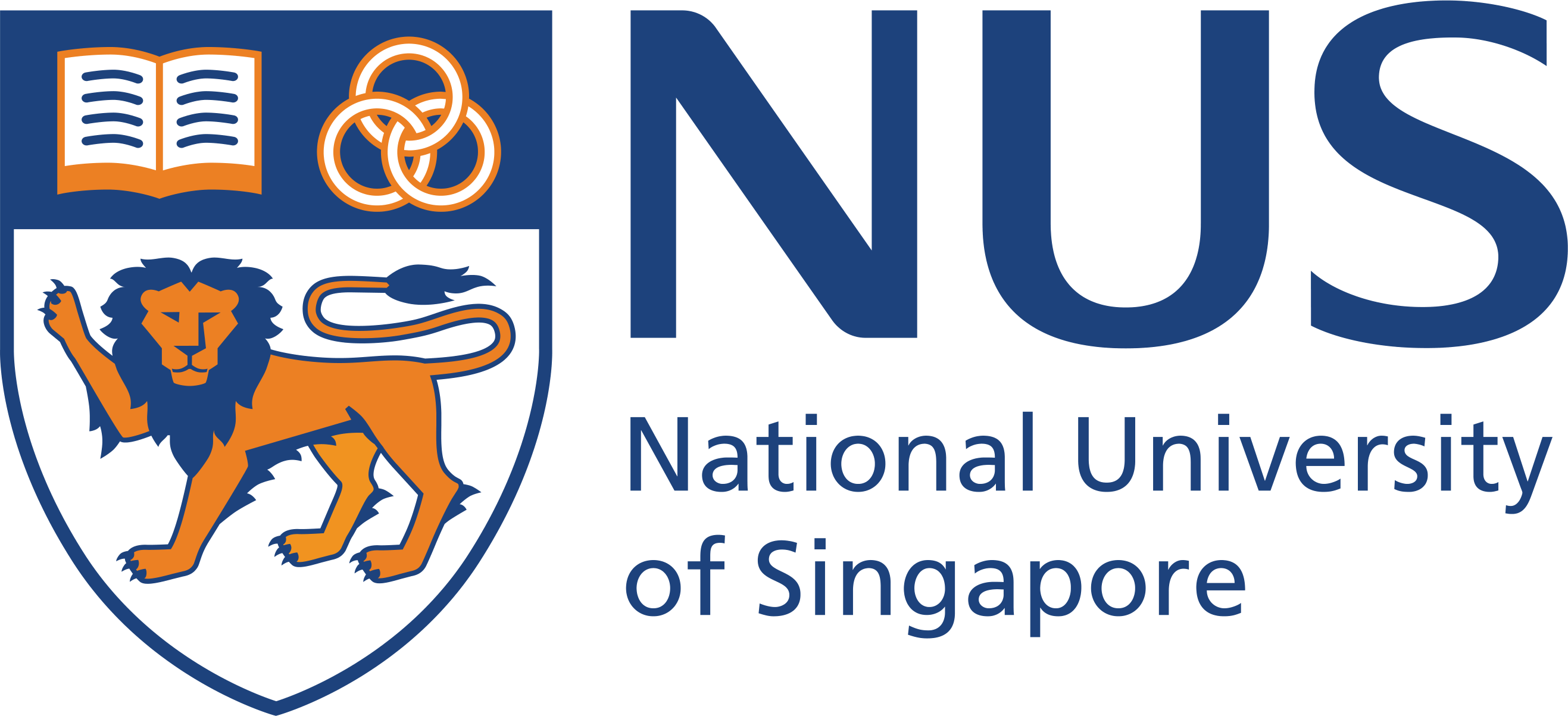}
\thirdleftlogo[5.8mm]{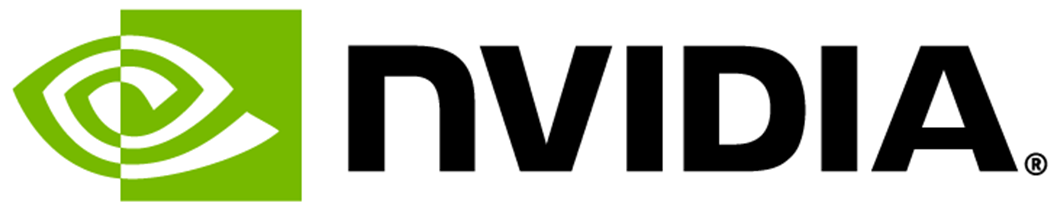}
\fourthleftlogo[8.2mm]{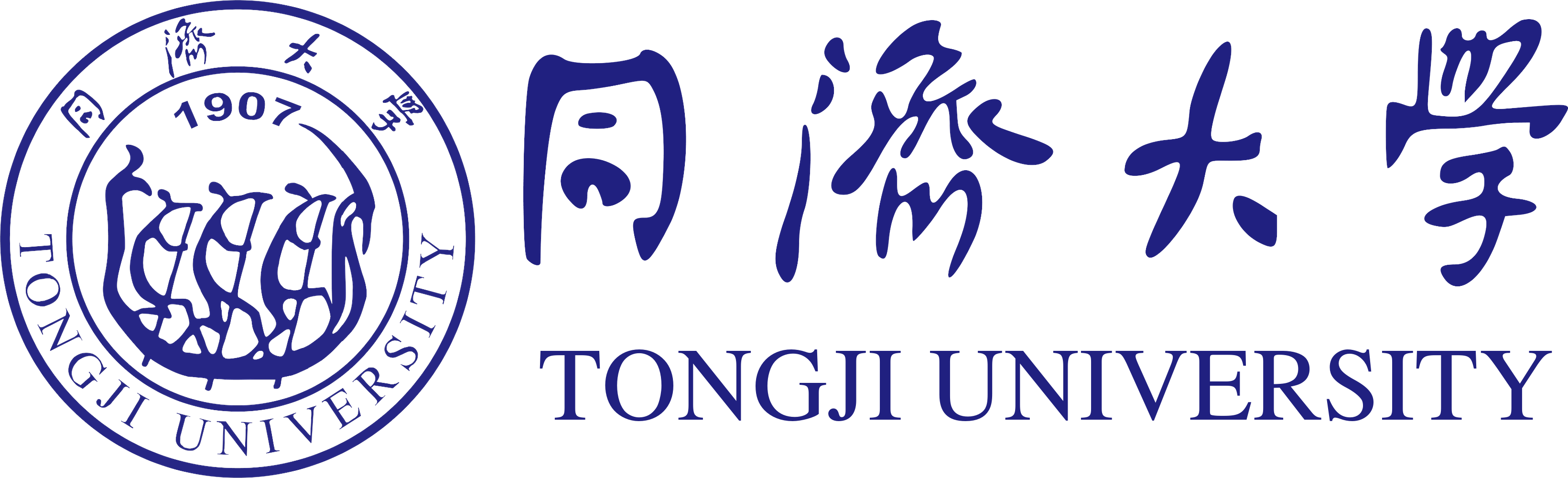}
\fifthleftlogo[6.4mm]{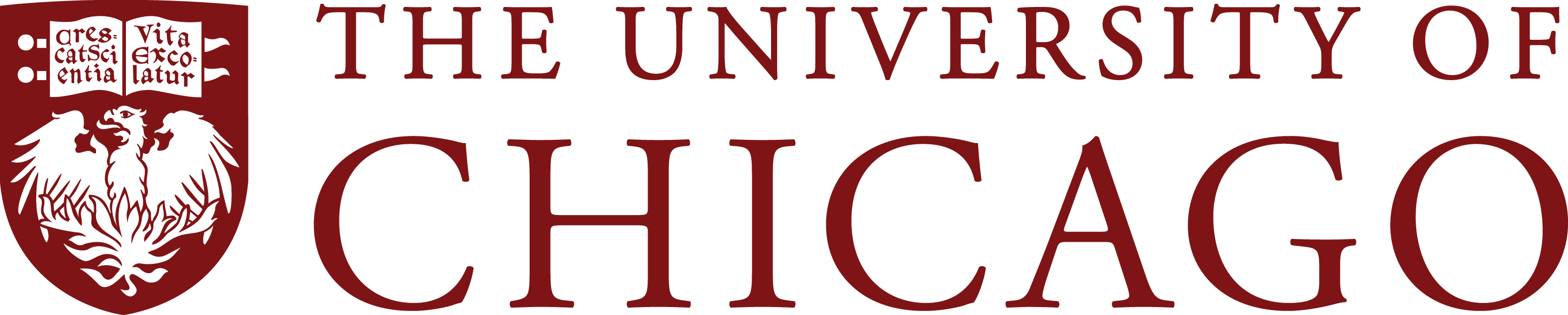}

\renewcommand{\authorfont}{\fontsize{11}{12.5}\selectfont}
\author[1,4*]{Feixiang Ruan}
\author[1*]{Guodong Lv}
\author[1*]{Kewei Zhang}
\author[1*]{Wenbo Huang}
\author[1]{Yintianrun Zhang}
\author[1]{Qiang Wang}
\author[1]{Ruiyang Si}
\author[1]{Shunyu Yang}
\author[1]{Shulong Jiang}
\author[4]{Weicheng Duan}
\author[1]{Rui Ni}
\author[1,2]{Jianxin Bi}
\author[1]{Lekai Wang}
\author[4]{Yuze Fan}
\author[4]{Haobo Li}
\author[4]{Zeliang Fu}
\author[5]{Jiankun Zhang}
\author[3]{Duomin Wang}
\author[3]{Enze Xie}
\author[1\dagger]{Daquan Zhou}
\affiliation[1]{Peking University}
\affiliation[2]{National University of Singapore}
\affiliation[3]{NVIDIA}
\affiliationnewline
\affiliation[4]{Tongji University}
\affiliation[5]{University of Chicago}
\contribution[*]{Equal contribution}
\contribution[\dagger]{Corresponding author}

\usepackage{fontawesome5}
\newcommand{\projectlink}[3]{\makebox[1.6em][l]{\normalsize #1}{\sffamily\bfseries\small #2:}~{\small\url{#3}}}
\abstract{Learning for dexterous manipulation is limited by the scarcity of high-quality demonstrations: retargeting from human video is imprecise, teleoperation gives no force feedback to the operator, and exoskeleton-based collection is hard to scale up.
We present \textbf{\ours}, a framework that enables coding agents to develop dexterous manipulation through \emph{programmatic policy}, without the need of demonstrations. The programmatic policy is developed in two stages: during \textit{exploration}, multiple agents explore in parallel for task skills and ways to compose them, and a distilling agent condenses the successful attempts into a task library and one program. During \textit{refinement}, this program is improved through randomized layouts for enhanced robustness and execution efficiency.
In simulated environments, \ours achieves SOTA on Bench2Dex with a score of 72.9 and reaches 81.0\% success on DexJoCo under full randomization, ahead of every trained policy on both benchmarks. In real world experiments on a bimanual platform with two 22-DoF hands, \ours learns to play piano, write calligraphy, and other tasks efficiently: learning a new piece from sheet music takes as little as 30 minutes. These results suggest that \ours is an effective framework for developing dexterous manipulation policies with coding agents in the absence of high quality dexterous manipulation demonstrations.
\par\vskip 3mm
\projectlink{\textcolor[HTML]{9A0000}{\faGlobeAmericas}}{Project Page}{https://vme0522.github.io/PyRUA-Dex/}\par\vskip 1.2mm
\projectlink{\faGithub}{GitHub Repo}{https://github.com/DAGroup-PKU/PyRUA-Lean/tree/dex}\par\vskip 1.2mm
\makebox[1.6em][l]{\normalsize\faEnvelope[regular]}{\sffamily\bfseries\small Corresponding Author:}~{\small\href{mailto:zhoudaquan21@gmail.com}{\texttt{Daquan Zhou}}}}

\date{\today}

\begin{document}
\maketitle

\begin{figure}[!ht]
  \centering
  \includegraphics[width=\textwidth]{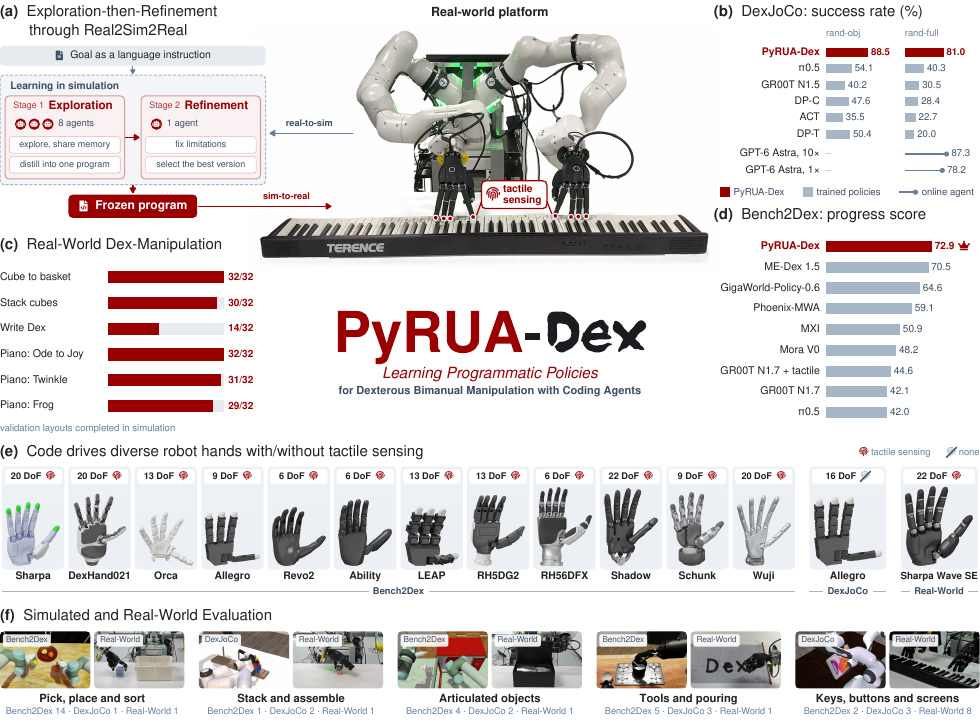}
  \caption[Overview]{\teasercaption}
  \vspace{-1em}
  \label{fig:overview}
\end{figure}

\section{Introduction}
\label{sec:introduction}

Dexterous manipulation is learned today mainly from demonstrations. Imitation
learning~\citep{zhao2023act,chi2023diffusionpolicy}, vision-language-action
models~\citep{brohan2023rt2,kim2024openvla,black2024pi0,black2025pi05,nvidia2025gr00t} and world-action
models~\citep{gigaai2026gigaworld} are all trained on teleoperated episodes, and data
generators~\citep{mandlekar2023mimicgen,jiang2025dexmimicgen} expand a few such episodes into many. For
multi-fingered hands, however, high-quality demonstrations are scarce. Retargeting human video to a robot hand is
imprecise, teleoperation gives the operator no force feedback during contact, and exoskeleton-based collection is
hard to scale. The data are also tied to the hand they were collected on, so a new hand requires new data and
retraining. Policies trained on such data remain fragile. On Bench2Dex, which evaluates bimanual tasks on 12
arm--hand embodiments~\citep{yang2026bench2dex}, the strongest trained policies succeed in little more than half of
the episodes~\citep{liauto2026medex,bench2dex2026leaderboard}, and all of them degrade when a layout's appearance or
object placement is perturbed.

Coding agents offer an alternative to collecting demonstrations. A \emph{programmatic policy} represents behavior
as a program rather than as network weights, so it can be read, tested and
verified~\citep{verma2018pirl,verma2019propel}. Language models already write programs that control
robots~\citep{liang2023codeaspolicies,singh2023progprompt,huang2023voxposer} and improve them from environment
feedback~\citep{wang2023voyager,ma2023eureka,shinn2023reflexion,wang2026pyruareal}, and general coding agents run,
test and debug software for hours~\citep{yang2024sweagent,openai2025codexcli}. Robot programs, however, are
typically written once from a prompt~\citep{liang2023codeaspolicies} or found by search in a small domain-specific
language~\citep{trivedi2021leaps,liu2023hprl}, and are not refined for robustness across layouts. For dexterous
hands, the agent has so far stayed in the loop instead: it observes and acts in every test
episode~\citep{sensenova2026astra}, consumes millions of tokens per episode, and still fails most episodes within
the benchmark's step budget. We ask whether a coding agent can \emph{learn} a dexterous task once, during
development, and produce a program that then runs on its own reliably without per-episode cost.

Learning such a program poses two distinct problems. The first is discovering a solution: the grasps, the ordering
of the two hands and the motions that complete each stage of the task. This is a search problem that requires many
attempts on a few layouts, and it proceeds faster when parallel agents share their progress. The second is making
the solution robust. A dexterous task consists of consecutive stages, and its success rate is the product of the
\emph{stage success rates}, the probability of completing each stage given that the preceding stages were completed.
A program that succeeds on one layout can therefore fail on most others: five stages at 80\% each give an overall
success rate of only 33\%. In our experiments, most failed runs end with steps to spare, at a stage that was not
completed (Section~\ref{sec:res-practice}), so failures come from unreliable stages rather than from the step
budget. Because the overall rate cannot exceed the lowest stage rate, robustness requires measuring every stage over
many layouts and improving the least reliable one.

\ours addresses these two problems in two stages (Figure~\ref{fig:method}). In the \emph{exploration} stage
(\stageone), multiple coding agents work on one task in parallel in simulation. They write and run code, rewind the
simulator to retry a stage, use the number of completed stages as feedback, and share their progress with one
another. A distilling agent then condenses the successful programs into a task library and one program. In the
\emph{refinement} stage (\stagetwo), an agent refines this program over many randomized layouts. It measures the
stage success rates, improves the least
reliable stage, and keeps a revision only if it performs no worse on layouts that no agent has seen. Agents and
programs control every robot through the same \emph{codebox}, a layer of hand, arm and sensing calls built from the
robot's description, so the procedure applies to any hand the codebox can model. The resulting policy is general
Python and runs frozen. It takes the inputs of an ordinary policy (calibrated cameras, joint positions and fingertip
touch), calls no language model, and runs at the robot's control rate. Because it measures the scene in every
episode, it remains reliable when a layout is perturbed.

On Bench2Dex, the frozen programs reach a score of 72.9, a new SOTA ahead of every trained policy, and at the
benchmark's step budget they succeed 2.4 times as often as the same model, GPT-6 Astra, acting online
(Sections~\ref{sec:res-main} and~\ref{sec:res-online}). On DexJoCo, a second benchmark with Allegro
hands~\citep{wang2026dexjoco}, they succeed in 81.0\% of episodes under full randomization, against 40.3\% for the
strongest trained policy, and are above each trained policy on every task (Section~\ref{sec:res-dexjoco}). On a real
bimanual platform with two 22-DoF hands, the same two stages run in a calibrated simulation that is updated with the
robot's feedback after each round, and the frozen programs then run on the robot (Section~\ref{sec:real}). They play
piano, write calligraphy and unpack objects, and learning a new piano piece from sheet music takes about 30 minutes.

In summary, our contributions are as follows:
\begin{itemize}
  \item \textbf{An agent-driven framework for dexterous manipulation.} We propose \ours, which enables coding agents
  to develop programmatic policies for dexterous manipulation without demonstrations. The resulting policy runs as
  frozen code at test time, with no trained model and no online agent.
  \item \textbf{Two-stage programmatic robot policy development using tactile feedback.} Parallel agents first
  explore in simulation for task skills and their composition. The skills are then refined over randomized layouts
  for robustness and efficiency, with fingertip tactile feedback used throughout.
  \item \textbf{Comprehensive simulated and real-world experiments across diverse robot hands and tasks.} \ours
  achieves SOTA on Bench2Dex. On a real bimanual platform with two 22-DoF hands, with system identification of the simulated hands, \ours develops code programs to perform piano playing, calligraphy and other tasks efficiently.
\end{itemize}
\section{Related Work}
\label{sec:related}

Our work lies at the intersection of dexterous manipulation, programmatic policies and coding for
robots.

\paragraph{Learning dexterous manipulation.}
Reinforcement learning has produced dexterous skills in simulation and on
hardware~\citep{openai2020dexterous,chen2022bidexhands}, at the price of task-specific rewards and large-scale
training per skill. Imitation learning from teleoperated demonstrations~\citep{zhao2023act,chi2023diffusionpolicy} and
vision-language-action models~\citep{brohan2023rt2,kim2024openvla,black2024pi0,black2025pi05,nvidia2025gr00t} are now
the dominant recipe, with data generators reducing the number of human
episodes~\citep{mandlekar2023mimicgen,jiang2025dexmimicgen}, and recent benchmarks evaluate them across hands and
settings~\citep{yang2026bench2dex,wang2026dexjoco,luo2026handson,wang2026efe,chen2025robotwin2}. To reach hardware,
a common route is to train in a simulation built from the deployment scene, with randomization covering what the
model gets wrong~\citep{torne2024rialto,chen2024urdformer,dai2024digitalcousins,tobin2017domain,peng2018sim,
wang2026r2seval}. \ours takes the same route to the real robot but trains no network: agents write and test a program
in simulation, and the program is what transfers.

\paragraph{Programmatic policies.}
Programmatic reinforcement learning represents a policy as a program in a symbolic language, which is easier to
interpret and verify than a neural policy~\citep{verma2018pirl,verma2019propel}; to keep the search over programs
tractable, the languages are small and the programs short~\citep{trivedi2021leaps,liu2023hprl}. Language models
write such programs directly: they compose perception and motion primitives from a task
description~\citep{liang2023codeaspolicies,singh2023progprompt,huang2023voxposer}, keep libraries of programs that
worked~\citep{wang2023voyager}, refine reward code from training outcomes~\citep{ma2023eureka} and generate tasks and
simulations~\citep{wang2024gensim,wang2024robogen}. These systems write a program once or improve it over a handful
of trials, mostly for parallel-jaw grippers. \ours writes general Python programs for multi-fingered hands and
develops them against success rates measured over hundreds of randomized layouts.

\paragraph{Coding agents for robots.}
A coding agent can control a robot directly, observing and issuing commands throughout every
episode~\citep{sensenova2026astra,su2026astra,robodojo2026leaderboard}; it keeps its full reasoning at test time but
pays for it in tokens, time and step budget in every episode. The alternative is to let agents develop programs that
then run without them. RAX scales exploration by coding agents to as many as 256 episodes per task and compiles what
it finds into modular programs~\citep{huang2026rax}, and the PyRUA family studies token-efficient
execution~\citep{si2026pyrualean}, data generation~\citep{zhang2026datafarm}, self-improvement on real
robots~\citep{wang2026pyruareal} and what agents learn from a benchmark's success
criteria~\citep{huang2026rsijudge}. \ours belongs to this second line and targets dexterous hands: parallel agents
explore a task in simulation, a second stage refines the resulting program stage by stage, and the program that is
handed over calls no language model.

\section{Method}
\label{sec:method}

\begin{figure*}[t]
  \centering
  \includegraphics[width=\textwidth]{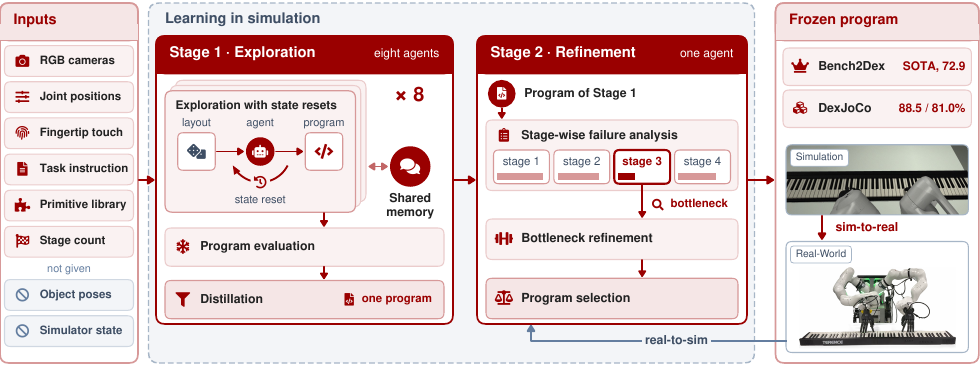}
  \caption{\textbf{Overview of \ours.} All learning takes place in simulation. \textbf{Stage 1 (exploration):} eight
    agents (one is drawn) explore the task, each on a training layout of its own with state resets, and share a
    memory; their programs are evaluated without the agents and distilled into a task library and one program.
    \textbf{Stage 2 (refinement):} one agent finds the program's bottleneck stage by a stage-wise failure analysis,
    refines it on randomized layouts and selects the version with the best held-out record. The resulting program
    runs frozen, without a language model. For the real world, the simulation is built from the real scene (real-to-sim), the
    frozen program is transferred (sim-to-real), and real-world feedback updates the simulation for the next round
    of refinement.}
  \label{fig:method}
\end{figure*}

\ours learns one programmatic policy per task in simulation and deploys it as a frozen program
(Figure~\ref{fig:method}). Learning proceeds in two stages, both carried out by general coding agents: Codex CLI
sessions~\citep{openai2025codexcli} of GPT-6 Astra~\citep{openai2026astra}, each equipped with a shell, its own
simulator instance and a library of perception and control primitives shared by all robots (the \emph{codebox}).
\emph{Exploration} (\stageone) searches for a program that completes the task at all; \emph{refinement} (\stagetwo)
improves how reliably that program completes the task on unseen layouts.

\subsection{Problem Formulation}
\label{sec:method-program}

For each task $\tau$, stated as a goal in natural language for two arms with multi-fingered hands, we seek a policy in
the form of a program $\pi_\tau$. At step $t$ it maps the observation history to an action,
\begin{equation}
a_t=\pi_\tau(o_{\le t}),
\label{eq:policy}
\end{equation}
where $o_t$ contains what a learned policy receives on the benchmark (the RGB images of the cameras with their
calibration, the joint positions of arms and hands, and the fingertip tactile maps) and $a_t$ is a joint-position
target. The program may call the robot's forward kinematics and a segmentation model~\citep{carion2025sam3}, but it
receives no simulator state, object pose, stage count or success signal, and it calls no language model.

A task is evaluated over \emph{layouts} $\xi\sim\mathcal{D}_\tau$, initial states of the scene that differ in object
poses, physical parameters and appearance, and its progress is measured by $K$ ordered \emph{stages}, sub-goals that
the simulator counts the first time they are reached. Let $\rho_k(\xi)\in\{0,1\}$ indicate that $\pi_\tau$ reaches
stages $1,\dots,k$ on layout $\xi$, with $\rho_0\equiv 1$, and let $\sigma(\xi)\in\{0,1\}$ indicate success, which
requires all stages and the task's final conditions within the episode's step budget, so that $\sigma\le\rho_K$. We
define the \emph{stage success rate} of stage $k$ as the fraction of layouts that reach it among those that reached
the stage before,
\begin{equation}
r_k(\pi_\tau)=\frac{\mathbb{E}_{\xi}\left[\rho_k(\xi)\right]}{\mathbb{E}_{\xi}\left[\rho_{k-1}(\xi)\right]},
\qquad k=1,\dots,K,
\label{eq:link}
\end{equation}
with the convention $0/0=0$, so that the success rate factorizes along the chain of stages:
\begin{equation}
\mathrm{SR}(\pi_\tau)=\mathbb{E}_{\xi}\left[\sigma(\xi)\right]=s(\pi_\tau)\prod_{k=1}^{K}r_k(\pi_\tau),
\qquad s(\pi_\tau)=\Pr\left[\sigma=1\mid\rho_K=1\right].
\label{eq:chain}
\end{equation}
The factorization in \eqref{eq:chain} motivates the two stages. A program is useful only once every stage success
rate is positive, which requires a search over grasps, hand orders and motions on a few layouts: the task of
exploration. Once every rate is positive, the product dominates, and because
$\partial\log\mathrm{SR}/\partial r_k=1/r_k$, the largest gain comes from the least reliable stage,
the \emph{bottleneck}, and finding it requires stage success rates estimated over many layouts: the task of
refinement.

Learning a task is therefore a search over programs under a development budget:
\begin{equation}
\max_{\pi_\tau\in\Pi}\ \mathrm{SR}(\pi_\tau)
\qquad\text{subject to}\qquad c_{\mathrm{LLM}}(\pi_\tau)=0,\quad n_{\mathrm{dev}}(\pi_\tau)\le B,
\label{eq:objective}
\end{equation}
where $\Pi$ is the set of programs of the form given in~\eqref{eq:policy}, $\mathrm{SR}$ the success rate on the
benchmark's test layouts, $c_{\mathrm{LLM}}$ the number of language-model calls in a test episode,
$n_{\mathrm{dev}}$ the number of simulated layouts spent on development and $B$ their budget
(Section~\ref{sec:setup}). The test layouts are never used during development. Agents interact only with
\emph{training layouts}, on which they also see the simulator's stage count, and a fixed set of 32 \emph{validation
layouts} per task, never shown to any agent, decides whether a new program replaces the current one.

\subsection{Policy Representation}
\label{sec:method-codebox}

\paragraph{Program structure.}
Every program has the same staged form: a \texttt{setup} function that measures the scene, and one function $f_k$
per stage, executed in order by a fixed runner within the episode's step budget (Appendix~\ref{sec:app-method},
Figure~\ref{fig:program}). Each $f_k$ starts from where $f_{k-1}$ left the robot and the scene, works toward stage
$k$ and returns whether, by the program's own measurement, the stage holds. A failed function does not abort the
episode: later functions check on entry what they need, and the runner calls the failed functions again while steps
remain. This form lets measurements be reported per stage, functions written by different agents be combined, and
a single stage be executed in isolation.

\paragraph{Perception and control primitives.}
Hands are what makes these tasks hard to program: the twelve hands of Bench2Dex have 6 to 22 driven joints, named and
arranged differently on each robot, and whether a grasp holds is revealed only by touch. Agents and programs therefore
control every robot through one library of primitives, which is generated from the robot's description file. It models
each hand as fingers and a thumb whose joints are grouped by function (bend, curl, spread), derives which object
widths each set of fingers can hold, and exposes hand primitives at that level: shape the hand, grasp an object of a
measured size, close until it is held, check that it is still held. Contact is read from the fingertip tactile maps
and from the lag between commanded and measured joint positions. Arm primitives move a frame of the hand to a target
and report an unreachable or colliding pose before moving, and perception primitives locate objects from the
calibrated cameras with SAM~3~\citep{carion2025sam3}. Every primitive returns an outcome with its reason, so a program
can check each step and retry. The same primitives drive all hands in our experiments
(Appendix~\ref{sec:app-method}).

\subsection{Stage 1: Exploration}
\label{sec:method-explore}

Exploration runs eight agents on a task in parallel. Each receives the task library, initialized from the task's
definition, a training layout of its own, and one of the stages not yet solved as its assignment.

\paragraph{Exploration with state resets.}
An agent controls the robot interactively through the primitives and queries the simulator's stage count after each
attempt. It can checkpoint the simulator's state during an episode and reset to a checkpoint, eight times per layout,
so that it retries one stage many times without repeating the stages before it.

\paragraph{Shared memory.}
Every attempt an agent finishes is
written to a memory shared by all agents of the task, with the stages it reached and the code that reached them. A
dedicated agent curates the memory, and each agent sees what is new in its next tool reply, so a stage solved by
one agent becomes the starting point of the others.

\paragraph{Program evaluation and distillation.}
Each agent ends by writing a program that reproduces what it achieved, and every program that reached a stage is
evaluated without the agent on eight fresh training layouts. A distilling agent then reads all evaluation records and
writes the task library: what is established about the task, how each stage was solved and how often, with code, and
what remains unsolved. It also selects the task's program: the evaluated program with the best record (completions,
then mean stage rate, then fewer steps), or a composition that takes each stage function from the agent whose program
reaches that stage most often, itself evaluated before it can be selected. The selected program replaces the current
one only if it is no worse on the validation layouts.

\subsection{Stage 2: Refinement}
\label{sec:method-refine}

Refinement starts from the program and task library of exploration, and a single agent edits the program.

\paragraph{Stage-wise failure analysis.}
Before the agent starts, the program runs without the agent on 32 fresh training layouts, and the result is broken
down by stage: for each stage, how many layouts entered it, how many reached it, how its function ended on the others
and how much of the step budget was left. Layouts are grouped by the first stage they missed, which gives the stage
success rates of~\eqref{eq:link} and shows the bottleneck before any change is made.

\paragraph{Bottleneck refinement.}
The agent works on one stage at a time, the bottleneck first. It opens failing layouts interactively, repairs the
stage function with the help of state resets, and then runs the program without itself in the loop on tens of fresh
layouts. To keep these measurements cheap, a \emph{partial rollout} executes the program only up to the stage under
work, so thirty layouts return in about the time those stages take on one. During refinement an episode has twice
the benchmark's step budget, so that a stage is made reliable before it is made fast.

\paragraph{Program selection.}
After every 160 training layouts, the current version is evaluated without the agent, at the benchmark's step budget,
on 32 fresh layouts and on a fixed held-out set of 32 layouts of which the agent sees only the counts. The agent stops
when the task completes on most held-out layouts, when the stage success rates stop improving, or when its 1,500
layouts are used up. A distilling agent selects the version with the best held-out record, and, as in exploration, this version
replaces the current program only if it is no worse on the validation layouts; otherwise the program from
exploration is kept.

\subsection{Sim-to-Real Transfer}
\label{sec:method-sim2real}

For a real robot, both stages run in a simulation of the robot's scene, and the robot itself only executes frozen
programs (Figure~\ref{fig:method}, right). We build the simulation from the platform's calibration and from objects
reconstructed from photographs (real-to-sim), randomize in every layout what the calibration leaves uncertain, and
run exploration once. Refinement then alternates with the robot: after each round the program is executed on the
robot, the discrepancies it reveals are written back into the simulation as updated physical parameters or task
rules, and the next round refines the same program in the updated simulation (Section~\ref{sec:real}).

\section{Experiments}
\label{sec:results}

We evaluate \ours on two simulated benchmarks for dexterous manipulation and, in Section~\ref{sec:real}, on a real
bimanual platform. The simulation experiments answer four questions, each about a separate comparison:
\begin{itemize}[leftmargin=1.4em,itemsep=1pt,topsep=3pt,parsep=0pt]
  \item \textbf{Q1:} Do programs written and iteratively refined by coding agents outperform policies trained on demonstrations?
  (Section~\ref{sec:res-main})
  \item \textbf{Q2:} What does refinement contribute beyond exploration? (Section~\ref{sec:res-practice})
  \item \textbf{Q3:} Is a frozen program better than the same model acting online? (Section~\ref{sec:res-online})
  \item \textbf{Q4:} Does the method generalize to another benchmark, with a different hand and simulator?
  (Section~\ref{sec:res-dexjoco})
\end{itemize}

\subsection{Experimental Setup}
\label{sec:setup}

\paragraph{Benchmarks and metrics.}
We use Bench2Dex~\citep{yang2026bench2dex}, with 26 bimanual tasks, each paired with one of 12 arm--hand
embodiments, and DexJoCo~\citep{wang2026dexjoco}, with 11 tasks for Allegro hands; together they cover twelve
dexterous hands with 6 to 22 driven joints (Figure~\ref{fig:overview}e,f). On Bench2Dex we follow the leaderboard
protocol~\citep{bench2dex2026leaderboard}: every task is evaluated on four channels of 50 episodes, the unperturbed
layouts (None) and layouts perturbed along the invariance axis, which leaves the correct action unchanged (Inv.),
along the equivariance axis, which changes it (Equi.), or both (Full). We report the success rate (SR), the latched
stage completion rate (LSCR, the mean fraction of stages reached), the efficiency (Eff, the expert's steps divided
by the steps of successful episodes) and the leaderboard's
$\mathrm{Score}=0.45\,\mathrm{SR}+0.45\,\mathrm{LSCR}+10\,\mathrm{Eff}$, with SR and LSCR in percent. On DexJoCo we
report the success rate over three seeds of 50 episodes per task, with the objects randomized (rand-obj) and with
cameras, lighting and textures randomized as well (rand-full). Appendix~\ref{sec:app-setup} describes both
benchmarks.

\paragraph{Implementation.}
All agents are Codex CLI sessions of GPT-6 Astra at the highest reasoning effort, with no memory across sessions.
Exploration uses eight agents per task, each with eight training layouts and eight state resets per layout.
Refinement uses one agent per task and at most 1,500 training layouts, with an evaluation on 32 fresh and 32 held-out
layouts after every 160 (the budget $B$ of~\eqref{eq:objective}). A program replaces the current one only if it is no worse on the 32 validation layouts. No
budget limits time or tokens. The final program of each task is evaluated once on the benchmark's test layouts, with
access to the program's own inputs only.

\paragraph{Baselines.}
We compare with two groups of methods (Appendix~\ref{sec:app-setup} lists them individually). \emph{Trained policies}
learn from the benchmarks' teleoperated demonstrations: on Bench2Dex, every leaderboard entry that reports all four
channels, covering imitation learning (IL), vision-language-action (VLA), world-action (WAM) and visuo-tactile action
(VTA) models~\citep{bench2dex2026leaderboard,gigaai2026gigaworld,phoenix2026mwa,liauto2026medex}; on DexJoCo, the
benchmark's five trained policies~\citep{wang2026dexjoco}. The \emph{online agent} is the GPT-6 Astra agent of
SenseNova~\citep{sensenova2026astra}, which uses the same model as \ours but controls the robot in every test episode
after preparing in context on the task's training data.

\begin{table*}[t]
\centering
\caption{\textbf{Bench2Dex results} (26 tasks; success rate per channel and four-channel means). Demos: trained on the
  benchmark's teleoperated demonstrations (ICL: read in context). LLM: a language model acts during evaluation.
  $^\dagger$Full channel only: the online agent's leaderboard entries, 260 episodes at 1$\times$ the step budget and 78
  at 10$\times$; the 78 episodes that its report runs at every budget give 25.6\% at 1$\times$, which
  Section~\ref{sec:res-online} uses.}
\label{tab:main}
\vspace{-0.4em}
\resizebox{\textwidth}{!}{%
\begin{tabular}{llcccccccccc}
\toprule
 & & & & \multicolumn{4}{c}{Success rate per channel (\%)} & \multicolumn{4}{c}{Four-channel mean} \\
\cmidrule(lr){5-8}\cmidrule(lr){9-12}
Method & Type & Demos & LLM & None & Equi. & Inv. & Full & SR (\%) & LSCR (\%) & Eff & Score \\
\midrule
ACT~\citep{zhao2023act} & IL & \cmark & \xmark & 29.3 & 15.5 & 14.8 & 12.8 & 18.1 & 34.9 & 0.967 & 33.5 \\
Diffusion Policy~\citep{chi2023diffusionpolicy} & IL & \cmark & \xmark & 8.9 & 3.0 & 3.2 & 3.4 & 4.6 & 18.3 & 0.865 & 19.0 \\
$\pi_{0.5}$~\citep{black2025pi05} & VLA & \cmark & \xmark & 35.2 & 21.2 & 26.8 & 26.2 & 27.4 & 44.7 & 0.960 & 42.0 \\
GR00T N1.5~\citep{nvidia2025gr00t} & VLA & \cmark & \xmark & 38.0 & 16.3 & 24.3 & 20.8 & 24.9 & 45.2 & 0.930 & 40.8 \\
GR00T N1.5 + tactile & VLA & \cmark & \xmark & 38.5 & 17.2 & 23.3 & 22.8 & 25.5 & 46.5 & 0.946 & 41.9 \\
GR00T N1.7 & VLA & \cmark & \xmark & 39.4 & 13.2 & 28.5 & 27.9 & 27.2 & 45.5 & 0.940 & 42.1 \\
GR00T N1.7 + tactile & VLA & \cmark & \xmark & 44.2 & 14.5 & 30.7 & 29.2 & 29.7 & 48.4 & 0.950 & 44.6 \\
Mora V0 & VLA & \cmark & \xmark & 38.8 & 29.5 & 32.5 & 32.1 & 33.2 & 52.3 & 0.974 & 48.2 \\
MXI & VLA & \cmark & \xmark & 38.4 & 37.3 & 35.8 & 36.0 & 36.9 & 54.8 & 0.970 & 50.9 \\
Phoenix-MWA~\citep{phoenix2026mwa} & WAM & \cmark & \xmark & 54.2 & 38.5 & 44.1 & 45.6 & 45.6 & 65.6 & 0.903 & 59.1 \\
GigaWorld-Policy-0.6~\citep{gigaai2026gigaworld} & WAM & \cmark & \xmark & 55.4 & 48.2 & 51.4 & 50.5 & 51.4 & 70.9 & 0.962 & 64.6 \\
ME-Dex 1.5~\citep{liauto2026medex} & VTA & \cmark & \xmark & 65.3 & 55.8 & 57.2 & 56.1 & 58.6 & 76.3 & 0.979 & 70.5 \\
\midrule
GPT-6 Astra agent, 1$\times$~\citep{sensenova2026astra} & Agent & ICL & \cmark & -- & -- & -- & 26.2$^\dagger$ & -- & -- & -- & -- \\
GPT-6 Astra agent, 10$\times$~\citep{sensenova2026astra} & Agent & ICL & \cmark & -- & -- & -- & 61.5$^\dagger$ & -- & -- & -- & -- \\
\midrule
\ours, exploration only & Agent & \xmark & \xmark & 44.9 & 45.8 & 44.3 & 43.4 & 44.6 & 67.0 & 0.889 & 59.1 \\
\rowcolor{RowRed} \ours & Agent & \xmark & \xmark & 60.8 & 62.1 & 58.0 & 61.5 & 60.6 & 82.3 & 0.859 & \textbf{72.9} \\
\bottomrule

\end{tabular}}
\vspace{-0.6em}
\end{table*}

\subsection{\ours Sets a New SOTA on Bench2Dex}
\label{sec:res-main}

\begin{figure*}[t]
  \centering
  \includegraphics[width=\textwidth]{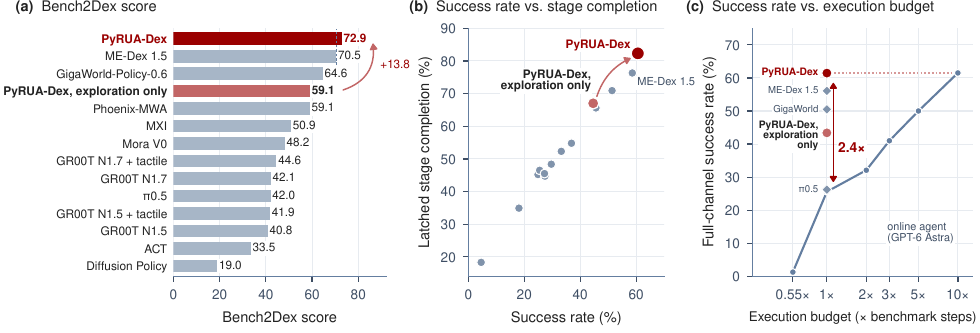}
  \caption{\textbf{Bench2Dex results.} \textbf{(a)}~Score of every four-channel entry; the dashed line marks the best
    trained policy. \textbf{(b)}~Success rate against latched stage completion of the same entries.
    \textbf{(c)}~Full-channel success rate against the execution budget; the frozen programs spend no tokens at test
    time; arrow: the benchmark's own budget; dotted line: where the online agent matches them.}
  \label{fig:results}
\end{figure*}

\ours scores 72.9 on Bench2Dex, a new SOTA (Table~\ref{tab:main}, Figure~\ref{fig:results}a). Its success rate of 60.6\% and stage completion of 82.3\% exceed those of
every trained policy, including ME-Dex 1.5, which fine-tunes one checkpoint per task. The lead is largest in stage
completion (Figure~\ref{fig:results}b): when a program fails, it has usually completed part of the task, because
each stage function checks its own result and the next one starts from what is there.

\paragraph{The programs are robust to perturbations.}
The success rate of \ours varies by only 4.1 points across the four channels, and the equivariance channel, in which
the perturbation changes the correct action, is its best. Every trained policy loses most on the perturbed
channels; GR00T N1.7 drops from 39.4\% to 13.2\%. A program measures the scene in every episode and computes its
motions from that measurement, whereas a trained policy reproduces the mapping it saw in the demonstrations.

\paragraph{The margin is widest where precision and order decide the task.}
\ours exceeds the best trained policy on \leadTwo{} of the 26 tasks (Appendix~\ref{sec:app-results},
Figure~\ref{fig:pertask}). The widest margins are on the bimanual piano melody (99.5\% against 32.0\%), ball sorting
and baking-tray preparation. A trained policy leads on the other 11 tasks. On three of them our programs almost never
succeed, each with a stage that no agent solved during development: placing the pan with the bread on its stand,
returning the bottle after pouring from hand to hand, and completing the jigsaw within the step budget. In answer to
Q1, programs written and refined by coding agents outperform policies trained on demonstrations on the benchmark as a
whole, but
not on every task.

\subsection{Refinement Makes Explored Programs Reliable}
\label{sec:res-practice}

\begin{figure*}[t]
  \centering
  \includegraphics[width=\textwidth]{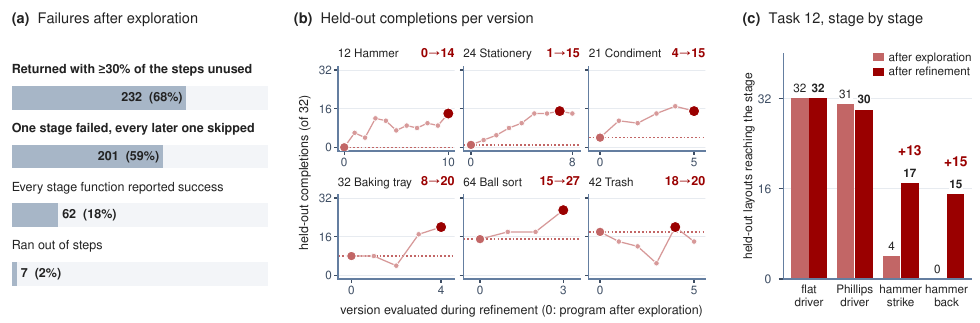}
  \caption{\textbf{Why refinement works.} \textbf{(a)}~How the 341 failed validation runs of the programs after
    exploration ended, on the 22 tasks that exploration completed at least once (a run can count in two classes).
    \textbf{(b)}~Held-out completions
    (of 32) of the versions evaluated during refinement; version 0 is the program after exploration, and the large
    dot is the version that was selected. \textbf{(c)}~Task 12: held-out layouts that reached each stage, after
    exploration and after refinement.}
  \label{fig:stage2}
\end{figure*}

Refinement raises the success rate from 44.6\% to 60.6\% and the score from 59.1 to 72.9, with gains on 20 of the 26
tasks (Appendix~\ref{sec:app-results}). The largest gains are on stationery sorting (6.5\% $\rightarrow$ 49.0\%),
baking-tray preparation (26.0\% $\rightarrow$ 68.0\%) and the hammer task, which exploration never completed (0\%
$\rightarrow$ 37.0\%).

\paragraph{Explored programs fail at a single stage, with budget to spare.}
We classify the 341 failed validation runs of the programs after exploration on the 22 tasks they completed at least
once (Figure~\ref{fig:stage2}a). In 68\% of them the program had returned with at least 30\% of the step budget
unused, and in 59\% one stage function had failed once and every later function had returned at once; only 2\% ran
out of steps. Calling failed stage functions again puts the unused budget to work, and refining the bottleneck
targets the stage that fails. On task 12 (Figure~\ref{fig:stage2}c), the program after exploration places both
screwdrivers on almost every held-out layout but strikes the wooden block on 4 of 32 and returns the hammer on none;
after refinement, on 17 and 15.

\paragraph{The gain carries over to unseen layouts.}
Held-out completions rise over successive versions on most tasks (Figure~\ref{fig:stage2}b). After exploration and
after refinement the programs complete 43.6\% and 61.7\% of the 832 validation layouts, close to their test success
rates of 44.6\% and 60.6\%. In answer to Q2, refinement adds 16.0 points by repairing the stage at which an explored
program fails, and the gain holds on unseen layouts.

\subsection{Frozen Programs Outperform the Same Model Acting Online}
\label{sec:res-online}

The online agent and \ours use the same model: the agent queries it in every test episode, \ours only during
development. At the benchmark's step budget, the frozen programs complete 61.5\% of the Full-channel episodes, 2.4
times the online agent's 25.6\% on the 78 episodes that its report runs at every
budget~\citep{sensenova2026astra} (Figure~\ref{fig:results}c). The online agent reaches 61.5\% only with ten times the
step budget, at 33.7M tokens per episode.

\paragraph{Development is paid once.}
Learning a task takes a median of 427M tokens and 21.6 agent hours in exploration and 214M tokens and 14.8 agent
hours in refinement (Appendix~\ref{sec:app-results}). Developing all 26 tasks took less than a third of the tokens
that the online agent spends on the benchmark's 5,200 test episodes at the step budget, and every further episode of
a frozen program costs no tokens and runs at the control rate. In answer to Q3, a frozen program is better: moving
the model's reasoning from test time to development yields both higher success and lower cost.

\subsection{The Method Generalizes to a Second Benchmark}
\label{sec:res-dexjoco}

\begin{table}[t]
\centering
\caption{\textbf{DexJoCo results} (success rate, \%; three seeds of 50 episodes per task and setting). Trained policies
  from the benchmark~\citep{wang2026dexjoco}.}
\label{tab:dexjoco}
\vspace{-0.4em}
\small
\begin{tabular}{llccrr}
\toprule
Method & Type & Demos & LLM & rand-obj & rand-full \\
\midrule
ACT~\citep{zhao2023act} & IL & \cmark & \xmark & 35.5 & 22.7 \\
DP-T~\citep{chi2023diffusionpolicy} & IL & \cmark & \xmark & 50.4 & 20.0 \\
DP-C~\citep{chi2023diffusionpolicy} & IL & \cmark & \xmark & 47.6 & 28.4 \\
$\pi_{0.5}$~\citep{black2025pi05} & VLA & \cmark & \xmark & 54.1 & 40.3 \\
GR00T N1.5~\citep{nvidia2025gr00t} & VLA & \cmark & \xmark & 40.2 & 30.5 \\
\midrule
GPT-6 Astra agent, 1$\times$~\citep{sensenova2026astra} & Agent & ICL & \cmark & -- & 78.2 \\
GPT-6 Astra agent, 10$\times$~\citep{sensenova2026astra} & Agent & ICL & \cmark & -- & 87.3 \\
\midrule
\ours, exploration only & Agent & \xmark & \xmark & 58.7 & 64.1 \\
\rowcolor{RowRed} \ours & Agent & \xmark & \xmark & 88.5 & 81.0 \\
\bottomrule

\end{tabular}
\vspace{-0.6em}
\end{table}

DexJoCo differs from Bench2Dex in its simulator (MuJoCo), its hand (the 16-DoF Allegro) and what it reveals during
development: it defines no stages, so an agent learns only whether an episode succeeded. \ours succeeds in 88.5\% of
the rand-obj episodes and 81.0\% of the rand-full ones, against 54.1\% and 40.3\% for the strongest trained policy,
$\pi_{0.5}$ (Table~\ref{tab:dexjoco}), and it exceeds the best trained policy on each of the 11 tasks in both settings
(Appendix~\ref{sec:app-results}, Figure~\ref{fig:dexjoco}). One program serves both settings and loses 7.5 points under full
randomization, whereas the trained policies, each trained for one setting, lose 9.7 to 30.4.

The two stages and the comparison with the online agent behave as on Bench2Dex: refinement adds 29.8 points under
rand-obj and 16.9 under rand-full, and at the benchmark's step budget the online agent succeeds in 78.2\% of the
rand-full episodes at 7.5M tokens per episode, against 81.0\% for the frozen programs at none, and needs twice the
budget to match them (Figure~\ref{fig:dexjoco}c). In answer to Q4, the method generalizes: unchanged, and with an
episode's success as its only feedback, it leads on a second benchmark with a different hand and simulator.

\section{Real-World Experiments}
\label{sec:real}

\begin{figure*}[t]
  \centering
  \includegraphics[width=\textwidth]{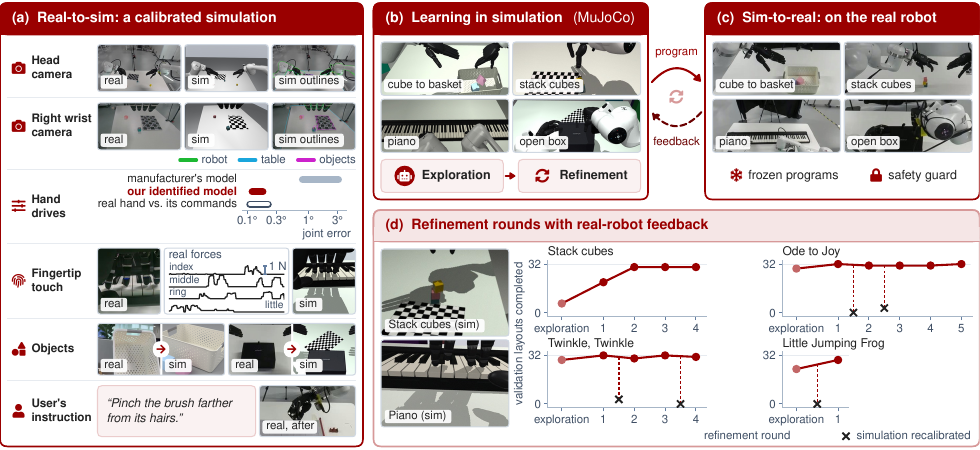}
  \caption{\textbf{Real-to-sim-to-real on our platform.} \textbf{(a)}~Real-to-sim: what the calibrated simulation
    shares with the real robot (cameras, the hands' drives, fingertip touch, objects and the user's instructions).
    \textbf{(b)}~Learning in simulation: exploration once, refinement in rounds. \textbf{(c)}~Sim-to-real: the frozen
    programs on the real robot, behind a safety guard; one pass from (b) to (c) and back is one round.
    \textbf{(d)}~Validation layouts
    completed in simulation (of 32) after exploration and after each round of refinement, one chart per task;
    $\times$: the program measured again after the simulation was recalibrated.}
  \label{fig:r2s2r}
\end{figure*}

We apply the same two stages to a real bimanual platform: two Franka FR3 arms with two Sharpa Wave SE hands (22
joints each), a ZED-M stereo camera above the table and an Intel RealSense D405 camera on each wrist
(Figure~\ref{fig:r2s2r}).

\paragraph{Real-to-sim: a calibrated simulation of the platform.}
The simulation is a MuJoCo~\citep{todorov2012mujoco} model built from the platform's calibration: the cameras'
intrinsics and poses, the hand-eye transforms of the wrist cameras, the table, the transform between the arms and
each hand's mount. Objects are reconstructed from photographs or a phone video and a few measured sizes. The hands
start from the manufacturer's model, and their drives are identified on the real hands under the controller the robot
uses (system identification, Appendix~\ref{sec:app-sysid}): on motions held out from the fit, the simulated joints
follow the real ones to 0.12--0.18$^\circ$, against 0.80--3.1$^\circ$ for the manufacturer's model. What the
calibration leaves uncertain, from the hands' gains and friction to camera poses, lighting and object masses, is
randomized in every layout (Appendix~\ref{sec:app-random}). A program receives the three RGB images with their calibration, the head camera's
depth, the joint positions and each fingertip's force and contact point, exactly as on the robot.

\begin{figure*}[t]
  \centering
  \includegraphics[width=0.92\textwidth]{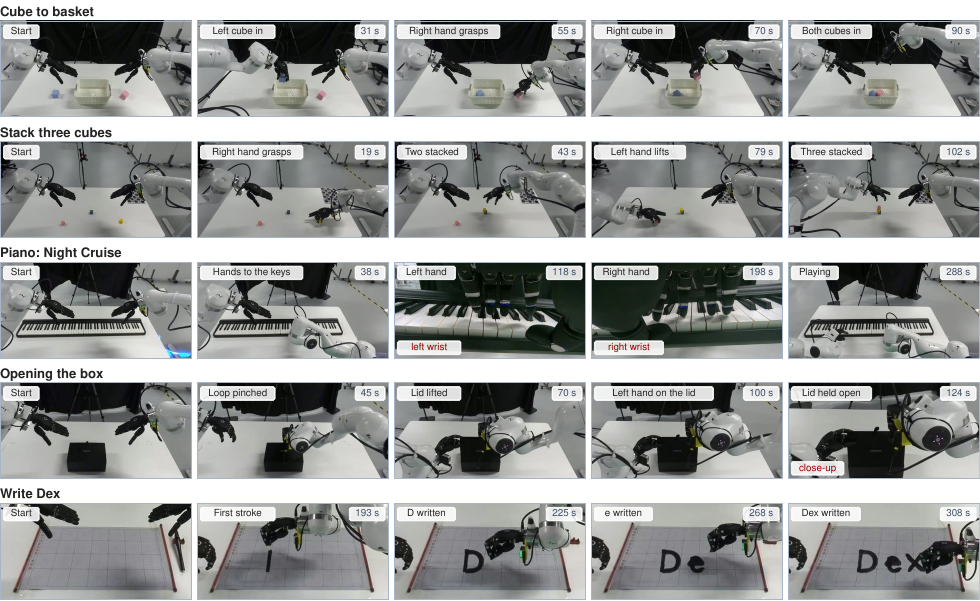}
  \caption{\textbf{Frozen programs on the real robot}, head camera unless marked: cube to basket, stacking three cubes,
    Night Cruise played with both hands, opening the box by its fabric loop, and writing ``Dex'' with a brush.}
  \label{fig:real}
\end{figure*}

\paragraph{Sim-to-real: refinement with real-robot feedback.}
Exploration runs once, as on the benchmarks, and refinement is repeated in rounds. After each round the program is
executed on the robot, where a safety guard rejects any motion that comes within a few millimeters of a joint
limit, the table or the other arm. What the robot reveals is written back
into the simulation, in its physics (the measured table height, the thumb's coupled swing, softer contacts) or in the
task's rules (a wrist camera kept clear of the table), together with the instructions the user gives at the robot,
and the next round starts in the updated simulation. Over these rounds the six tasks rise from 119 to 168 of their
192 validation layouts (Table~\ref{tab:realtasks}, Figure~\ref{fig:r2s2r}d); stacking three cubes improves from 6 to 20
and then to 30. When an update of the simulation breaks a program, the next round restores it: Ode to Joy goes from
32 to 0 and back to 31. On the robot, the frozen programs put both cubes into the basket, build the tower of three cubes, open the box by its fabric
loop and hold it open, take the brush and write ``Dex'', and play Night Cruise on the piano with both hands
(Figure~\ref{fig:real}).

\begin{table}[t]
\centering
\caption{\textbf{Tasks learned for the real robot.} Validation layouts completed in simulation (of 32) after
  exploration and after the last round of refinement. Steps at 30~Hz.}
\label{tab:realtasks}
\vspace{-0.4em}
\small\setlength{\tabcolsep}{4.6pt}
\begin{tabular}{llrrrr}
\toprule
Task & Goal & Steps & Exploration & Refinement & Rounds \\
\midrule
Cube to basket & each hand puts its cube in the basket & 1200 & 29 & 32 & 1 \\
Stack cubes & a tower of three cubes & 1000 & 6 & 30 & 4 \\
Write Dex & take the brush and write ``Dex'' on the cloth & 1500 & 3 & 14 & 1 \\
Ode to Joy & piano, the right hand, eight phrases & 4500 & 29 & 32 & 5 \\
Twinkle, Twinkle & piano, two hands, six phrases & 4500 & 29 & 31 & 4 \\
Little Jumping Frog & piano, two hands, eight phrases & 6000 & 23 & 29 & 1 \\
\midrule
Total & & & 119/192 & 168/192 & \\
\bottomrule
\end{tabular}
\vspace{-0.6em}
\end{table}

\paragraph{A skill learned once transfers within a task family.}
The piano tasks share one skill, a player distilled from the learned programs that arranges a score for two hands
and plays it. With this skill a new piece typically takes 30 minutes to an hour of iteration, whereas the first
pieces took exploration and up to five rounds of refinement. Jasmine, Night Cruise, Secret Base, the whole Little
Jumping Frog and Canon were produced this way.

\section{Limitations and Future Work}
\label{sec:limitations}

\paragraph{Limitations.}
Coding agents write motion well and sustained contact poorly. The programs are strongest where geometry, order and
timing decide a task, and weakest where continuous control of force and slip is required: trained policies lead on
the frying pan, the pour from hand to hand and the jigsaw puzzle, and assembly is the lowest of the DexJoCo tasks
(34.0\% under object randomization). A program acts through explicit targets and thresholds, and the simulated
fingertip sensors render contact coarsely (the tactile maps of Bench2Dex respond only on flat faces square to the
pad), so the programs infer contact mainly from joint lag. Refinement strengthens stages that already hold on some
layouts; a stage that never holds remains a problem of exploration. Finally, each task is learned separately, and on
the real robot we show the frozen programs running but measure their reliability in the calibrated simulation rather
than over repeated hardware trials.

\paragraph{Future work.}
The most direct extension is to let a program hand its contact phases to learned controllers and to develop it
against a more faithful tactile simulation. A second direction is transfer across tasks: a task library stores what
worked, with code and measured rates, and the piano pieces show that a skill distilled from it carries over within a
family; seeding the exploration of a new task with the libraries of related tasks would extend this across families. A third direction is a systematic evaluation on hardware, with success rates over
repeated trials on the real robot.

\clearpage
\bibliographystyle{plainnat}
\setlength{\bibhang}{0pt}
\setlength\bibindent{0pt}
\bibliography{references}

\newpage
\appendix
\section{Benchmarks, Baselines and Evaluation}
\label{sec:app-setup}

\paragraph{Bench2Dex.}
Bench2Dex~\citep{yang2026bench2dex} runs in Isaac Lab and pairs each of its 26 bimanual tasks with one of 12 arm--hand
embodiments: UR5, xArm7, Franka Panda, KUKA IIWA7, RM65 and JAKA ZU7 arms with Shadow, Allegro, LEAP, Ability, Orca,
Schunk, Wuji, Sharpa, Revo2, DexHand021, RH56DFX and RH5DG2 hands, with 6 to 22 driven joints per hand. The tasks
include tool use (hammer and screwdrivers, a ladle for soup), articulated objects (microwave, fridge, toilet lid,
faucet), pouring, sorting and multi-stage packing, and a bimanual piano melody. Each scene has five cameras, among them
a head camera and one camera on each wrist, and tactile maps on the fingertip pads. We evaluate under the
leaderboard's protocol at the benchmark commit used for submission, with the benchmark's own result files checked by
its validator (5,200 episodes, no errors). Efficiency is averaged over the tasks with at least one success, and SR,
LSCR and Eff are averaged equally over channels and tasks.

\paragraph{DexJoCo.}
DexJoCo~\citep{wang2026dexjoco} runs in MuJoCo and has 11 tasks for Franka arms with 16-DoF Allegro hands, six for
one arm (hammering a nail, clicking a mouse, picking up a bucket, pinching tongs, folding glasses, watering a plant)
and five for two (unlocking an iPad, the Tower of Hanoi, assembly, cooking with a microwave, taking a photograph).
Its rand-obj setting randomizes the placement of the objects and the height of the table; rand-full also randomizes
the camera poses, the lighting and the tabletop texture. A program receives the benchmark's policy observation, the
RGB images of its cameras without calibration, the flange poses and the hand joints, and commands flange poses and
hand joint targets at 30~Hz. Both stages run in rand-full. Refinement ran on ten tasks, and the photograph task keeps
its program from exploration. One frozen program per task is evaluated in both settings under the official protocol,
three seeds of 50 episodes per task and setting (1,650 episodes per setting).

\paragraph{Baselines.}
On Bench2Dex the trained policies are imitation learning (ACT~\citep{zhao2023act}, Diffusion
Policy~\citep{chi2023diffusionpolicy}), vision-language-action models ($\pi_{0.5}$~\citep{black2025pi05}, GR00T
N1.5~\citep{nvidia2025gr00t} and N1.7 with and without tactile input, Mora V0, MXI), world-action models
(Phoenix-MWA~\citep{phoenix2026mwa}, GigaWorld-Policy-0.6~\citep{gigaai2026gigaworld}) and ME-Dex 1.5, a
visuo-tactile action model with task-specific checkpoints~\citep{liauto2026medex}; all of them train on the
benchmark's 1.3K teleoperated demonstrations. The online GPT-6 Astra agent prepares in context on each task's
training data and then controls the robot in every test episode; it reports the Full channel only. Its report runs
78 episodes (three per task) at budgets from 0.55$\times$ to 10$\times$ the step budget and gives its token
costs~\citep{sensenova2026astra}; we compare at equal budget and cost on those episodes, and
Table~\ref{tab:main} lists its leaderboard entries (260 episodes at 1$\times$, 78 at 10$\times$). On DexJoCo the
trained policies are ACT, Diffusion Policy with transformer and CNN backbones (DP-T, DP-C), $\pi_{0.5}$ and GR00T
N1.5, each trained per task and setting on 100 teleoperated demonstrations~\citep{wang2026dexjoco}, and the online
agent runs at 0.5--10$\times$ the step budget in rand-full~\citep{sensenova2026astra}. RAX also reports DexJoCo
results~\citep{huang2026rax}, with a different model (Claude Opus 5.5 in Claude Code); as the model differs, they are
not comparable, and we leave them out.

\paragraph{Agents and compute.}
Exploration ran eight agents per task (208 sessions), one agent that curates the shared memory and one distilling
agent; refinement ran one agent per task and one distilling agent. Each agent is one conversation. Simulators run on
NVIDIA B200 nodes with eight GPUs each, one to four simulators per GPU, and every simulator shares its GPU with a
SAM~3 worker; three refinement runs used an RTX PRO 6000 cluster with the same code.

\paragraph{Test evaluation.}
The test evaluation runs the final program under the benchmark's protocol with access to the program's own calls only
(observation, stepping, calibration, tactile maps, forward kinematics and segmentation), without stage count and
without a language model. The test layouts decide nothing: a task whose refinement produced no better program keeps
the test record of its program from exploration.

\FloatBarrier
\section{Method Details}
\label{sec:app-method}

\begin{algorithm}[!htbp]
\caption{\ours for one task. $\pi^\ast$: the current program; $\mathrm{Val}(\hat\pi)\succeq\mathrm{Val}(\pi^\ast)$: no
  fewer completions on the 32 validation layouts, then no lower mean stage rate.}
\label{alg:method}
\vspace{-0.4em}
\begin{algobox}
\small
\begin{minipage}[t]{0.49\linewidth}
\algstage{Stage 1: Exploration}\\[-0.9em]
\begin{tabbing}
\hspace{1.4em}\=\hspace{1.0em}\=\hspace{1.0em}\=\kill
1:\> $\mathcal{L}\gets$ library initialized from the task's definition\\
2:\> \algkw{for} agents $i=1,\dots,8$ \algkw{in parallel do}\\
3:\>\> explore a training layout with state resets\\
4:\>\> add every attempt to the shared memory\\
5:\>\> $\pi_i\gets$ program that reproduces the attempt\\
6:\>\> evaluate $\pi_i$ without the agent on 8 fresh layouts\\
7:\> $(\mathcal{L},\hat\pi)\gets$ \textsc{Distill}(evaluation records)\\
8:\> $\pi^\ast\gets\hat\pi$ \algkw{if} $\mathrm{Val}(\hat\pi)\succeq\mathrm{Val}(\pi^\ast)$
\end{tabbing}
\end{minipage}\hfill
\begin{minipage}[t]{0.49\linewidth}
\algstage{Stage 2: Refinement}\\[-0.9em]
\begin{tabbing}
\hspace{1.6em}\=\hspace{1.0em}\=\hspace{1.0em}\=\kill
9:\> analyze $\pi^\ast$ per stage on 32 fresh layouts\\
10:\> \algkw{while} layouts remain \algkw{and not} submitted \algkw{do}\\
11:\>\> $k\gets$ bottleneck stage; repair $f_k$ with state resets\\
12:\>\> partial rollouts of stage $k$ on tens of layouts\\
13:\>\> every 160 layouts: 32 fresh + 32 held-out\\
14:\> $\hat\pi\gets$ best version on the held-out set, else $\pi^\ast$\\
15:\> $\pi^\ast\gets\hat\pi$ \algkw{if} $\mathrm{Val}(\hat\pi)\succeq\mathrm{Val}(\pi^\ast)$\\
16:\> \algkw{return} $\pi^\ast$ \algcmt{runs frozen, no language model}
\end{tabbing}
\end{minipage}
\end{algobox}
\vspace{-0.8em}
\end{algorithm}

\begin{figure}[!htbp]
\begin{minipage}[t]{0.497\linewidth}
\algstage{A program, written by the agents for one task}
\begin{lstlisting}[style=prog]
STAGES = ["grasp", "place"]  # the task's stages, in order
AGAIN = True     # call stages that did not end ok again

def setup(codebox, context):
    # measure once what the stages need
    context["table"] = codebox.table()
    context["box"] = codebox.locate("box")

def grasp(codebox, context):
    tool = codebox.locate("screwdriver")       # measure
    codebox.move_to("right", grasp_point(tool))   # act
    codebox.close("right")
    if not codebox.holding("right"):           # check
        return Outcome(False, "not held")
    return Outcome(True, "held")

def place(codebox, context):
    if not codebox.holding("right"):    # check on entry
        return Outcome(False, "nothing held")
    codebox.move_to("right", above(context["box"]))
    codebox.release("right")
    seen = codebox.relocate("screwdriver")   # measure again
    return Outcome(inside(seen, context["box"]), "in box")
\end{lstlisting}
\end{minipage}\hfill
\begin{minipage}[t]{0.497\linewidth}
\algstage{The runner, the same for every task}
\begin{lstlisting}[style=prog]
def run_episode(program, robot):
    codebox, context = Codebox(robot), {}
    program.setup(codebox, context)
    todo, last = program.STAGES, {}
    while todo and codebox.steps_left() > 0:
        used = codebox.steps_used
        for name in todo:                  # in order
            stage = getattr(program, name)
            last[name] = stage(codebox, context)
            # a failed stage does not stop the ones after it
        # one pass, unless AGAIN and the pass used steps
        if not program.AGAIN or codebox.steps_used == used:
            break
        todo = [n for n in program.STAGES if not last[n].ok]
        context["again"] = last    # how each last ended
\end{lstlisting}
\algstage{An episode's log (example)}
\begin{lstlisting}[style=prog,language={}]
[stage] setup begins at step 0
[stage] grasp: not held (not ok)
[stage] place: nothing held (not ok)
[stage] pass 2: again for grasp, place
[stage] grasp: held (ok)
[stage] place: in box (ok)
\end{lstlisting}
\end{minipage}
\vspace{-0.4em}
\caption{\textbf{The form of every program.} Left: the template every program fills in, here with two stages; the
  agents' programs fill each stage with many more calls (the final program of task 12 has 2,241 lines in 22 files).
  Right: the runner, the same for every task, condensed from its code. It calls \texttt{setup}, then the stage
  functions in order; a failed stage does not stop the ones after it, and with \texttt{AGAIN = True} the stage
  functions that did not end ok are called again while steps remain and a pass still uses steps. Below the runner,
  an example of the log it writes for one episode: the first call of \texttt{grasp} fails, \texttt{place} returns at
  once, and the second pass completes both.}
\label{fig:program}
\end{figure}

\paragraph{Program and runner.}
A programmatic policy for a task with stages $1,\dots,K$ consists of a \texttt{setup} function and one stage function
$f_k$ per stage (Figure~\ref{fig:program}). \texttt{setup} measures the scene into a context; $f_k$ starts where
$f_{k-1}$ left robot and scene and returns an outcome that says whether, by the program's own measurement, the stage
holds. With \texttt{AGAIN = True} the runner makes further passes, in which every function that did not end ok is
called again, in order, while steps remain and a pass still uses steps; each function is told how it last ended. A
program reads its remaining step budget from the codebox and never assumes a number.

\paragraph{A model of every hand.}
The codebox builds a model of each hand once, from the robot's public description alone: the kinematic tree, the
commanded and coupled joints with their ranges, the links' collision shapes and the links that carry the fingertip
pads. It reads no link or joint name. Each fingertip link makes a finger, and the thumb is the finger whose base stands
out of the row of the others. A finger's joints are grouped by what they do: the joints that bend the finger at its
base, those that curl the rest of it, and those that spread it sideways, each group on one scale from open (0) to
closed (1); the thumb can also oppose and flex. Which way a joint closes is read from its range.

\paragraph{Grasps instead of joint angles.}
From the model the codebox works out, for every set of fingers, which object widths their pads can hold squarely (the
pads within $30^\circ$ of facing each other) or at an angle (up to $45^\circ$), and which bars the palm can wrap. A
program asks to grasp an object it has measured, and the codebox picks the fingers and how far to open them. The same
calls (\texttt{shape}, \texttt{close} with a squeeze, \texttt{grasp}, \texttt{release} and single-finger motions)
drive the hands of all 12 Bench2Dex robots and the Sharpa hands of our real platform.

\paragraph{Touch.}
A program does not assume that a grasp held; it checks. \texttt{holding} reads the fingertips' touch and compares how
far the hand closed with how far it closes when empty, and \texttt{feel} moves a hand until something is touched. Touch
comes from the fingertip maps and from joint lag, the gap between a joint's commanded and measured position: a finger
that meets an object falls behind its command. Each robot's drive responses (lag, ramps, settling) were measured once
in the simulator, so this lag is told apart from the lag of a free motion.

\paragraph{Arms.}
Arm calls put a frame of the hand (the grasp point, the palm, a fingertip or a pad) at a point, with an approach
direction, via points and a hand shape formed on the way. \texttt{arc} moves the hand on a circle about an axis, for
doors, lids and knobs, and \texttt{together} runs one call per arm at the same time. Every call plans whole before it
moves: it solves all its poses first and, without having moved, returns \texttt{ik\_unreachable} with the nearest pose
it can take, \texttt{table} when a link would go below the measured table, or \texttt{needs\_reconfigure} with the cost
of reconfiguring the arm, which it takes only when allowed. All motions run at the joints' speed limits, so a program
saves steps by needing fewer motions, not by moving faster.

\paragraph{Perception.}
Sensing uses only the episode's camera calibration and the robot's own sensors. \texttt{segment} returns
SAM~3~\citep{carion2025sam3} masks for a ladder of prompts, from an object's name to its appearance; \texttt{locate}
combines an object's masks in the calibrated fixed cameras into its position, footprint, height and yaw, at no cost in
steps; \texttt{relocate} and \texttt{look} measure it again with a wrist camera, from where the hand is or from close
above; \texttt{table} and \texttt{probe} measure the table's height, from the cameras and by a light touch-down. Every
call returns an outcome with its reason, data and the steps it took, and never raises for a failure it names.

\paragraph{From an episode to a program.}
An agent explores interactively, and what it found must become a program that runs alone. When an explored layout
that reached a stage is closed, the harness compiles the path the agent kept into a \emph{recipe}: the palm's
waypoints relative to the objects the agent located, split into phases at every closing and opening of a hand, with
the grasps it made. It also writes a draft program in the staged form whose stage functions play the phases that
led to each stage. A recipe is open loop: it carries free motion and targets relative to objects, and contact does not
carry over by itself. The agent turns the draft into a program by closing the loop: measuring again before contact,
checking that a hand holds, and stating how each stage function fails.

\paragraph{Exploration.}
A state reset returns the episode to a checkpoint by running it again from its start, and no simulated time
passes while an agent thinks. An agent's session has eight fresh training layouts, one to explore and seven on which
its program then runs whole without the agent. The shared memory holds single episodes with state resets: its entries
say that a method exists, never how often it works, and a program copies what it uses rather than depending on them.
Only what earned a stage enters the task library, and every rate in it comes from evaluations without the agent. The
library has
fixed sections: the task and what is established about it; what scored; how each stage scored and where it failed,
with method, code and rate as $n$ of $N$ fresh layouts; the checks these runs depend on; the step budget; and what is
not solved yet, from which the next assignments are drawn.

\paragraph{Refinement.}
A partial rollout runs the program up to and including the function of a chosen stage, holds the robot for 3\,s of
simulated time so that a released object comes to rest, and ends the episode there; its record is the simulator's own
count. Every reply to the agent reports the stages: for each, the layouts that entered and reached it, how the
function ended on the others, the steps of a call, how often the function was called again and with what result, and
whether the program's own verdict agreed with the simulator's count. Every episode the agent starts and every layout
of a rollout counts toward its 1,500 training layouts; the periodic evaluations draw their own layouts, and further
work is refused until the evaluation that is due has been run. Between versions that differ by a layout or two on the
held-out set, the agent keeps the one whose stages hold more often. The submitted program and the current one also run
on 32 fresh layouts that nobody has used, which gives an unbiased rate. On the benchmarks refinement runs once per
task; for the real robot it is repeated in rounds (Section~\ref{sec:real}).

\paragraph{What an agent is given.}
Every agent of either stage receives the same material and nothing learned elsewhere: the primitives' manual (the
calls of the codebox, their arguments and how each reports its outcome); the robot's structure as the hand model
derives it (fingers, joints, limits, grasp frames) and the measured joint responses; the benchmark's rules (what an
episode is, what a program may read, the step limit); and the task's goal in words, written by hand from the task's
definition, with every stage stated in terms an agent can see and measure. The goal texts never describe the
benchmark's checking code and give no way to satisfy a stage other than doing it. While developing, an agent reads
the simulator's stage count after each episode; a program never does. Exploration starts from an empty task library;
refinement starts from the library and program that exploration left, and from nothing that the refinement of any
other task produced.

\FloatBarrier
\section{Additional Results}
\label{sec:app-results}

\paragraph{Failure modes of the online agent.}
The online agent's report names the two failure modes that refinement targets: a failed prerequisite that blocks
every later subtask, and progress that a later motion undoes~\citep{sensenova2026astra}. Its leaderboard entry of
260 episodes at the step budget has 26.2\%, against 25.6\% on the 78 episodes of its report.

\paragraph{Cost.}
Exploration ran 208 agent sessions, \tokOne{}B tokens and \hrsOne{} agent hours over the 26 tasks, a median of 427M
tokens and 21.6 agent hours per task. Refinement ran \tokTwo{}B tokens and \hrsTwo{} agent hours over the 23 tasks
with recorded counts, a median of 214M tokens and 14.8 agent hours per task. The cost of refinement follows the
difficulty of the task: nearly solved tasks such as the cleaner box and the piano melody were submitted after one
evaluation and about four agent hours, while the hardest, such as the frying pan and the wine pour from hand to hand,
used all or nearly all 1,500 layouts and up to 38 agent hours. Refinement for all 26 tasks ran in about three days of
wall-clock time on shared GPU clusters.

\begin{table}[!htbp]
\centering
\caption{\textbf{Exploration versus refinement} on the same 26 Bench2Dex tasks. Agent cost of refinement over the 23
  tasks run on our B200 cluster.}
\label{tab:stages}
\vspace{-0.4em}
\small
\begin{tabular}{lcc}
\toprule
 & After exploration & After refinement \\
\midrule
Agents per task & 8 in parallel & 1 \\
Validation layouts completed (26 tasks $\times$ 32) & 363/832 (43.6\%) & 513/832 (61.7\%) \\
Test success rate (\%) & 44.6 & 60.6 \\
Test latched stage completion (\%) & 67.0 & 82.3 \\
Efficiency (tasks with a success) & 0.889 (22) & 0.859 (24) \\
Bench2Dex score & 59.1 & \textbf{72.9} \\
Tasks with no success & 4 & 2 \\
Tasks above the best trained policy & 8/26 & 15/26 \\
Agent tokens per task, median (M) & 427 & 214 \\
Agent hours per task, median & 21.6 & 14.8 \\
Layouts used for development & 8 per agent, 8 per agent-free replay & 245--1500 per task \\
\bottomrule

\end{tabular}
\end{table}

\begin{figure*}[!htbp]
  \centering
  \includegraphics[width=\textwidth]{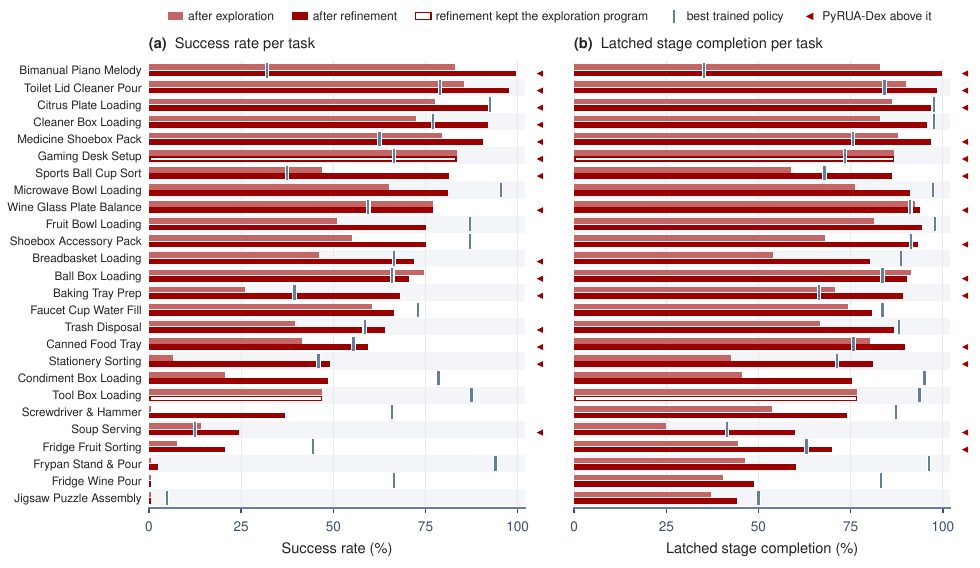}
  \caption{\textbf{Per-task results on Bench2Dex.} \textbf{(a)}~Success rate and \textbf{(b)}~latched stage completion
    on each task: the program after exploration (light bar) above the program after refinement (dark bar), and the
    best trained policy on the task (tick). Hollow bar: refinement kept the program from exploration. Triangles:
    \ours above the best trained policy (\leadTwo{} of 26 tasks in success rate, 14 in stage completion).}
  \label{fig:pertask}
\end{figure*}

\begin{table*}[!htbp]
\centering
\caption{\textbf{Per-task results on Bench2Dex.} SR and LSCR: four-channel means (\%). Val.: completions on the 32
  validation layouts. Layouts, evaluations and tokens (M): the refinement of the task (``--'': counts not kept).
  $^\ast$Refinement kept the program from exploration.}
\label{tab:pertask}
\vspace{-0.4em}
\resizebox{\textwidth}{!}{%
\begin{tabular}{rllrrrrrrrrr}
\toprule
 & & & \multicolumn{2}{c}{SR} & \multicolumn{2}{c}{LSCR} & \multicolumn{2}{c}{Val. (of 32)} & \multicolumn{3}{c}{Refinement} \\
\cmidrule(lr){4-5}\cmidrule(lr){6-7}\cmidrule(lr){8-9}\cmidrule(lr){10-12}
ID & Task & Embodiment & Expl. & Ref. & Expl. & Ref. & Expl. & Ref. & Layouts & Evaluations & Tokens \\
\midrule
03 & Wine Glass Plate Balance & xArm7 + Ability & 77.0 & 77.0 & 92.5 & 93.7 & 22 & 26 & 372 & 4 & 175 \\
06 & Fruit Bowl Loading & UR5 + RH56DFX & 51.0 & 75.0 & 81.4 & 94.2 & 17 & 21 & 358 & 2 & 115 \\
07 & Citrus Plate Loading & UR5 + RH5DG2 & 77.5 & 93.0 & 86.1 & 97.8 & 20 & 30 & 412 & 3 & 125 \\
08 & Frypan Stand \& Pour & UR5 + Schunk & 0.0 & 2.5 & 46.2 & 60.2 & 0 & 0 & 1500 & 12 & 823 \\
09 & Cleaner Box Loading & xArm7 + LEAP & 72.5 & 92.0 & 83.0 & 95.8 & 16 & 27 & 245 & 1 & 58 \\
12 & Screwdriver Box \& Hammer & UR5 + RH56DFX & 0.0 & 37.0 & 53.6 & 74.1 & 0 & 12 & 1500 & 10 & 309 \\
21 & Condiment Box Loading & UR5 + Wuji & 20.5 & 48.5 & 45.5 & 75.3 & 5 & 17 & 745 & 5 & 306 \\
22 & Tool Box Loading & Panda + Orca & 47.0 & 47.0$^\ast$ & 76.8 & 76.8 & 15 & 15 & 1408 & 9 & 462 \\
24 & Stationery Category Sorting & RM65 + Revo2 & 6.5 & 49.0 & 42.6 & 81.0 & 10 & 13 & 1486 & 8 & 684 \\
26 & Canned Food Tray Arrangement & IIWA7 + Sharpa & 41.5 & 59.5 & 80.2 & 89.6 & 12 & 26 & 352 & 3 & 84 \\
27 & Ball Box Loading & UR5 + Wuji & 74.5 & 70.5 & 91.4 & 90.4 & 21 & 21 & -- & -- & -- \\
32 & Baking Tray Prep & IIWA7 + Sharpa & 26.0 & 68.0 & 70.9 & 89.1 & 10 & 18 & 506 & 4 & 153 \\
34 & Fridge Wine Interhand Pour & UR5 + RH5DG2 & 0.0 & 0.0 & 40.3 & 48.7 & 0 & 0 & 1500 & 10 & 751 \\
42 & Trash Disposal & UR5 + RH56DFX & 39.5 & 64.0 & 66.7 & 86.7 & 10 & 26 & 717 & 5 & 250 \\
43 & Fridge Fruit Shelf Sorting & UR5 + Shadow & 7.5 & 20.5 & 44.4 & 69.9 & 5 & 11 & -- & -- & -- \\
44 & Microwave Bowl Loading & UR5 + Schunk & 65.0 & 81.0 & 76.1 & 91.0 & 19 & 25 & 503 & 4 & 169 \\
51 & Toilet Lid Cleaner Pour & xArm7 + LEAP & 85.5 & 97.5 & 90.0 & 98.3 & 28 & 32 & 641 & 3 & 207 \\
60 & Breadbasket Fast-Food Loading & UR5 + RH5DG2 & 46.0 & 72.0 & 53.9 & 80.2 & 15 & 21 & -- & -- & -- \\
61 & Medicine Shoebox Pack & Panda + Orca & 79.5 & 90.5 & 87.8 & 96.8 & 27 & 27 & 687 & 5 & 292 \\
62 & Shoebox Accessory Pack & xArm7 + LEAP & 55.0 & 75.0 & 68.2 & 93.2 & 14 & 23 & 354 & 2 & 72 \\
64 & Sports Ball Cup Sort & Panda + Allegro & 47.0 & 81.5 & 58.8 & 86.2 & 19 & 25 & 314 & 3 & 76 \\
67 & Faucet Cup Water Fill & RM65 + Revo2 & 60.5 & 66.5 & 74.4 & 80.8 & 17 & 28 & 499 & 3 & 214 \\
73 & Jigsaw Puzzle Assembly & IIWA7 + Sharpa & 0.0 & 0.0 & 37.0 & 44.1 & 0 & 0 & 852 & 6 & 298 \\
76 & Soup Serving & UR5 + Shadow & 14.0 & 24.5 & 25.0 & 59.9 & 5 & 9 & 1500 & 9 & 660 \\
79 & Bimanual Piano Melody & xArm7 + Ability & 83.0 & 99.5 & 83.0 & 99.8 & 27 & 31 & 245 & 1 & 59 \\
80 & Gaming Desk Setup & JAKA ZU7 + DexHand021 & 83.5 & 83.5$^\ast$ & 86.8 & 86.8 & 29 & 29 & 587 & 3 & 241 \\
\bottomrule

\end{tabular}}
\end{table*}

\begin{figure*}[!htbp]
  \centering
  \includegraphics[width=\textwidth]{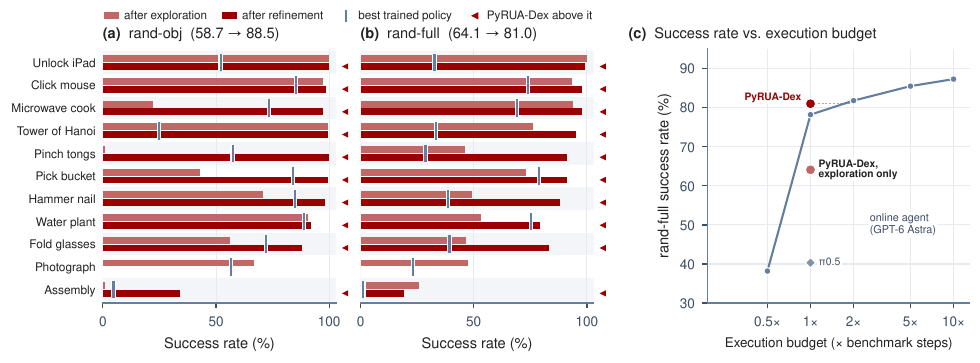}
  \caption{\textbf{DexJoCo results per task.} \textbf{(a)}~rand-obj and \textbf{(b)}~rand-full success rate: the
    program after exploration (light bar) above the program after refinement (dark bar), with the averages in the
    panel titles, and the best trained policy on the task (tick). Triangles: \ours above it, on all ten tasks
    that were refined; the photograph task was not refined and has one bar. \textbf{(c)}~rand-full success rate against
    the execution budget; the frozen programs spend no tokens at test time, the dotted line marks where the online
    agent matches them, and the diamond is the best trained policy, $\pi_{0.5}$.}
  \label{fig:dexjoco}
\end{figure*}

\begin{table*}[!htbp]
\centering
\caption{\textbf{DexJoCo per task} (success rate, \%, mean $\pm$ std over three seeds). Expl., Ref.: \ours after
  exploration and after refinement. Trained: the best of the
  benchmark's five trained policies~\citep{wang2026dexjoco}. Bold: above it. $^\ast$Not refined. $^\dagger$With a
  fallback to the fixed camera when the randomized one is absent; without it, 0.0 under rand-obj (rand-obj averages
  49.8 and 79.5).}
\label{tab:dexjoco-pertask}
\vspace{-0.4em}
\small
\begin{tabular}{lcrrrrrr}
\toprule
 & & \multicolumn{3}{c}{rand-obj} & \multicolumn{3}{c}{rand-full} \\
\cmidrule(lr){3-5}\cmidrule(lr){6-8}
Task & Arms & Expl. & Ref. & Trained & Expl. & Ref. & Trained \\
\midrule
Hammer nail & 1 & 70.7$\,\pm\,$3.1 & \textbf{98.0$\,\pm\,$3.5} & 84.7 ($\pi_{0.5}$) & 49.3$\,\pm\,$6.4 & \textbf{88.0$\,\pm\,$6.9} & 38.7 (GR00T) \\
Click mouse$^\dagger$ & 1 & 97.3$\,\pm\,$1.2 & \textbf{98.7$\,\pm\,$2.3} & 85.3 (GR00T) & 93.3$\,\pm\,$1.2 & \textbf{98.0$\,\pm\,$2.0} & 74.0 (GR00T) \\
Pick bucket & 1 & 42.7$\,\pm\,$9.2 & \textbf{99.3$\,\pm\,$1.2} & 84.0 ($\pi_{0.5}$) & 73.3$\,\pm\,$5.0 & \textbf{91.3$\,\pm\,$1.2} & 78.7 ($\pi_{0.5}$) \\
Pinch tongs & 1 & 0.0$\,\pm\,$0.0 & \textbf{100.0$\,\pm\,$0.0} & 57.3 (DP-C) & 46.0$\,\pm\,$3.5 & \textbf{91.3$\,\pm\,$3.1} & 28.7 (DP-C) \\
Fold glasses & 1 & 56.0$\,\pm\,$8.0 & \textbf{88.0$\,\pm\,$10.6} & 72.0 ($\pi_{0.5}$) & 46.7$\,\pm\,$5.8 & \textbf{83.3$\,\pm\,$11.7} & 39.3 ($\pi_{0.5}$) \\
Water plant & 1 & 90.7$\,\pm\,$1.2 & \textbf{92.0$\,\pm\,$2.0} & 88.7 ($\pi_{0.5}$) & 53.3$\,\pm\,$4.2 & \textbf{79.3$\,\pm\,$7.6} & 75.3 ($\pi_{0.5}$) \\
Unlock iPad & 2 & 100.0$\,\pm\,$0.0 & \textbf{100.0$\,\pm\,$0.0} & 52.0 (DP-C) & 100.0$\,\pm\,$0.0 & \textbf{99.3$\,\pm\,$1.2} & 32.7 ($\pi_{0.5}$) \\
Tower of Hanoi & 2 & 99.3$\,\pm\,$1.2 & \textbf{99.3$\,\pm\,$1.2} & 24.7 (DP-T) & 76.0$\,\pm\,$6.9 & \textbf{95.3$\,\pm\,$1.2} & 33.3 ($\pi_{0.5}$) \\
Assembly & 2 & 0.0$\,\pm\,$0.0 & \textbf{34.0$\,\pm\,$2.0} & 4.7 (DP-T) & 26.0$\,\pm\,$2.0 & \textbf{19.3$\,\pm\,$9.0} & 1.3 (GR00T) \\
Microwave cook & 2 & 22.0$\,\pm\,$13.1 & \textbf{97.3$\,\pm\,$2.3} & 73.3 (DP-T) & 94.0$\,\pm\,$5.3 & \textbf{98.0$\,\pm\,$3.5} & 69.3 ($\pi_{0.5}$) \\
Photograph$^\ast$ & 2 & 66.7$\,\pm\,$4.2 & \textbf{66.7$\,\pm\,$4.2} & 56.7 (DP-T) & 47.3$\,\pm\,$3.1 & \textbf{47.3$\,\pm\,$3.1} & 23.3 ($\pi_{0.5}$) \\
\midrule
Average &  & 58.7 & \textbf{88.5} & 54.1 ($\pi_{0.5}$) & 64.1 & \textbf{81.0} & 40.3 ($\pi_{0.5}$) \\
\bottomrule

\end{tabular}
\end{table*}

\FloatBarrier
\section{Real-Robot Details}
\label{sec:app-real}

\subsection{System Identification of the Hands}
\label{sec:app-sysid}

System identification estimates the parameters of a model from measurements of the real system, so that the model
responds to a command as the real system does. We apply it to the drives of the hands, because the programs depend
on them: the codebox reads contact from the lag between a joint's commanded and measured position
(Appendix~\ref{sec:app-method}), so a simulated finger that follows its commands differently from the real one would
report contact where there is none, or miss it, once the program runs on the robot. The manufacturer's model
describes the hand's geometry, but its drive parameters do not match the hand as we run it: on the test motions
below, its joints deviate from the real ones by 0.80--3.1$^\circ$ (Table~\ref{tab:sysid}).

\paragraph{What is identified.}
We keep the links, joints and meshes of the manufacturer's MJCF model of the Sharpa Wave SE and set, for each of the
22 joints of a hand, three groups of drive parameters. The first is the controller. On the robot, a joint with angle
$q$ and reference $q^{\mathrm{ref}}$ is driven with the torque
\begin{equation}
\tau=K_p\left(q^{\mathrm{ref}}-q\right)+K_d\left(\dot q^{\mathrm{ref}}-\dot q\right),
\label{eq:pd}
\end{equation}
where $K_p$ is the stiffness and $K_d$ the damping. Our controller raises the factory gains (1.5 to 2 times the
stiffness, up to 1.7 times the damping), adds an 80\,Hz dither scheduled by each joint's reference speed, and
corrects the reference in advance. In simulation, each joint has a position actuator with the same gains, whose
torque is $K_p(u-q)-K_d\,\dot q$ for a command $u$; commanding $u=q^{\mathrm{ref}}+(K_d/K_p)\,\dot q^{\mathrm{ref}}$
reproduces the torque of the robot's controller exactly. The second group is the joint's range and its speed and
torque limits, which we read from the real hand (the model's 0--15$^\circ$ for the little finger's CMC joint is
0--25$^\circ$ on the hand). The third group is the joint's friction loss $f$ (its dry friction) and its armature $a$
(the inertia of the rotor as seen at the joint), which we fit to the real joint's response.

\paragraph{Fitting.}
For joint $j$ we record the real joint while it follows single-joint reference motions at 10, 50 and
150$^\circ$/s, and replay the same references in simulation. With $\mathcal{M}_j$ the set of these recordings and
$T_m$ the number of samples of recording $m$, the parameters $\theta_j=(f_j,a_j)$ minimize the mean squared
difference between the simulated and the measured joint angle,
\begin{equation}
\theta_j^\ast=\arg\min_{\theta_j}\;\frac{1}{|\mathcal{M}_j|}\sum_{m\in\mathcal{M}_j}\frac{1}{T_m}\sum_{t=1}^{T_m}
\left(q^{\mathrm{sim}}_{j,m}(t;\theta_j)-q^{\mathrm{real}}_{j,m}(t)\right)^2 ,
\label{eq:sysid}
\end{equation}
where $q^{\mathrm{real}}_{j,m}(t)$ is the joint's measured angle at sample $t$ of recording $m$ and
$q^{\mathrm{sim}}_{j,m}(t;\theta_j)$ its simulated angle under the parameters $\theta_j$. We solve this problem with
the Nelder--Mead method, one joint at a time, for the 44 joints of the two hands.

\paragraph{Validation.}
We test the identified model on motions that the fit did not use: one joint moving slowly, five-finger grasps at
three speeds, a finger wave and random smooth motions of all joints (Table~\ref{tab:sysid}). The simulated joints
follow the real ones to 0.12--0.18$^\circ$, against 0.80--3.1$^\circ$ for the manufacturer's model. The real hand
follows its own references to 0.11--0.22$^\circ$ on the same motions, so the remaining difference between
simulation and robot is as small as the hand's own tracking error.

\begin{table}[!htbp]
\centering
\caption{\textbf{System identification of the hands}, on motions not used in the fit: root-mean-square difference of
  joint angles in degrees, median over the joints that move, for the left (L) and right (R) hand. Manufacturer's and
  identified: simulation against the real hand. Real hand: the real hand against its own reference.}
\label{tab:sysid}
\vspace{-0.4em}
\small
\begin{tabular}{lrcccccc}
\toprule
 & & \multicolumn{2}{c}{Manufacturer's model} & \multicolumn{2}{c}{Identified model} & \multicolumn{2}{c}{Real hand} \\
\cmidrule(lr){3-4}\cmidrule(lr){5-6}\cmidrule(lr){7-8}
Motion & Joints & L & R & L & R & L & R \\
\midrule
One joint at 2$^\circ$/s & 6 & 0.80 & 0.81 & 0.14 & 0.12 & 0.13 & 0.11 \\
Five-finger grasp, slow & 16 & 1.02 & 1.05 & 0.13 & 0.13 & 0.13 & 0.13 \\
Five-finger grasp, medium & 16 & 1.53 & 1.58 & 0.15 & 0.14 & 0.15 & 0.15 \\
Five-finger grasp, fast & 16 & 3.07 & 3.09 & 0.17 & 0.18 & 0.22 & 0.21 \\
Finger wave & 8 & 2.34 & 2.37 & 0.13 & 0.13 & 0.15 & 0.14 \\
Random smooth motion & 22 & 0.99 & 1.00 & 0.16 & 0.16 & 0.16 & 0.16 \\
\bottomrule

\end{tabular}
\end{table}

\paragraph{What is not identified.}
The dither is not simulated. Contact stiffness is not identified either: the simulation reports each fingertip's
force and contact point as the robot does, but their scale is not calibrated against the robot's sensors. For the
arms, the simulation uses the FR3's own limits and the same trajectory generator as the robot. What identification
leaves uncertain is randomized, as described next.

\subsection{Randomization and Objects}
\label{sec:app-random}

\paragraph{Randomization.}
Every layout of the simulation draws what the calibration leaves uncertain: the hands' gains, friction and armature,
the pads' friction, the cameras' poses within their calibration accuracy, the arms' bases, the cameras' noise and
color, the lighting, distractor objects, and the objects' masses and frictions.

\paragraph{Cube to basket.}
Cube to basket is the running example of Figure~\ref{fig:r2s2r}: its objects were reconstructed from a phone video,
and every layout draws the cubes' edge, the basket's size, all masses and frictions, and the placement of basket and
cubes.

\FloatBarrier
\section{Reproducing the Figures and Tables}
\label{sec:app-repro}

Every figure is drawn by a script in \texttt{figures/<name>/make\_fig.py} from the data in \texttt{figures/\_data/},
and every table of results by \texttt{scripts/make\_tables.py}; each figure folder lists the source of every picture and number
in its \texttt{SOURCES.md}. The Bench2Dex numbers are those of our leaderboard submissions, whose per-episode logs
are public with the leaderboard; the DexJoCo numbers are the summaries of the official evaluation, kept with the data.

\end{document}